\documentclass[sigconf]{acmart}

\AtBeginDocument{%
  }

\setcopyright{none}
\renewcommand\footnotetextcopyrightpermission[1]{}

\usepackage{pifont}
\usepackage{multirow}
\usepackage{booktabs}
\usepackage{subcaption}
\begin{document}

\title{GS-STVSR: Ultra-Efficient Continuous Spatio-Temporal Video Super-Resolution via 2D Gaussian Splatting}

\author{
    Mingyu Shi$^{1\dagger}$, \quad 
    Xin Di$^{1\dagger, \ddagger}$, \quad 
    Long Peng$^{1}$, \quad 
    Boxiang Cao$^{1}$, \quad 
    Anran Wu$^{1}$, \quad 
    Zhanfeng Feng$^{1}$, \\
    Jiaming Guo$^{2}$, \quad
    Renjing Pei$^{2}$, \quad
    Xueyang Fu$^{1}$, \quad
    Yang Cao$^{1*}$, \quad
    Zhengjun Zha$^{1*}$ \\[0.2cm]
    $^1$University of Science and Technology of China \quad
    $^2$Huawei Noah's Ark Lab \\[0.1cm]
    {\tt\small \{shimy2003, dx9826\}@mail.ustc.edu.cn, forrest@ustc.edu.cn, zhazj@ustc.edu.cn}
}

\renewcommand{\shortauthors}{Shi et al.}

\begin{abstract}
Continuous Spatio-Temporal Video Super-Resolution (C-STVSR) aims to simultaneously enhance the spatial resolution and frame rate of videos by arbitrary scale factors, offering greater flexibility than fixed-scale methods that are constrained by predefined upsampling ratios. In recent years, methods based on Implicit Neural Representations (INR) have made significant progress in C-STVSR by learning continuous mappings from spatio-temporal coordinates to pixel values. However, these methods fundamentally rely on dense pixel-wise grid queries, causing computational cost to scale linearly with the number of interpolated frames and severely limiting inference efficiency. We propose GS-STVSR, an ultra-efficient C-STVSR framework based on 2D Gaussian Splatting (2D-GS) that drives the spatiotemporal evolution of Gaussian kernels through continuous motion modeling, bypassing dense grid queries entirely. We exploit the strong temporal stability of covariance parameters for lightweight intermediate fitting, design an optical flow-guided motion module to derive Gaussian position and color at arbitrary time steps, introduce a Covariance resampling alignment module to prevent covariance drift, and propose an adaptive offset window for large-scale motion. Extensive experiments on Vid4, GoPro, and Adobe240 show that GS-STVSR achieves state-of-the-art quality across all benchmarks. Moreover, its inference time remains nearly constant at conventional temporal scales ($\times 2$--$\times 8$) and delivers over $3\times$ speedup at extreme scales ($\times 32$), demonstrating strong practical applicability.
\end{abstract}


\begin{teaserfigure}
  \includegraphics[width=\textwidth]{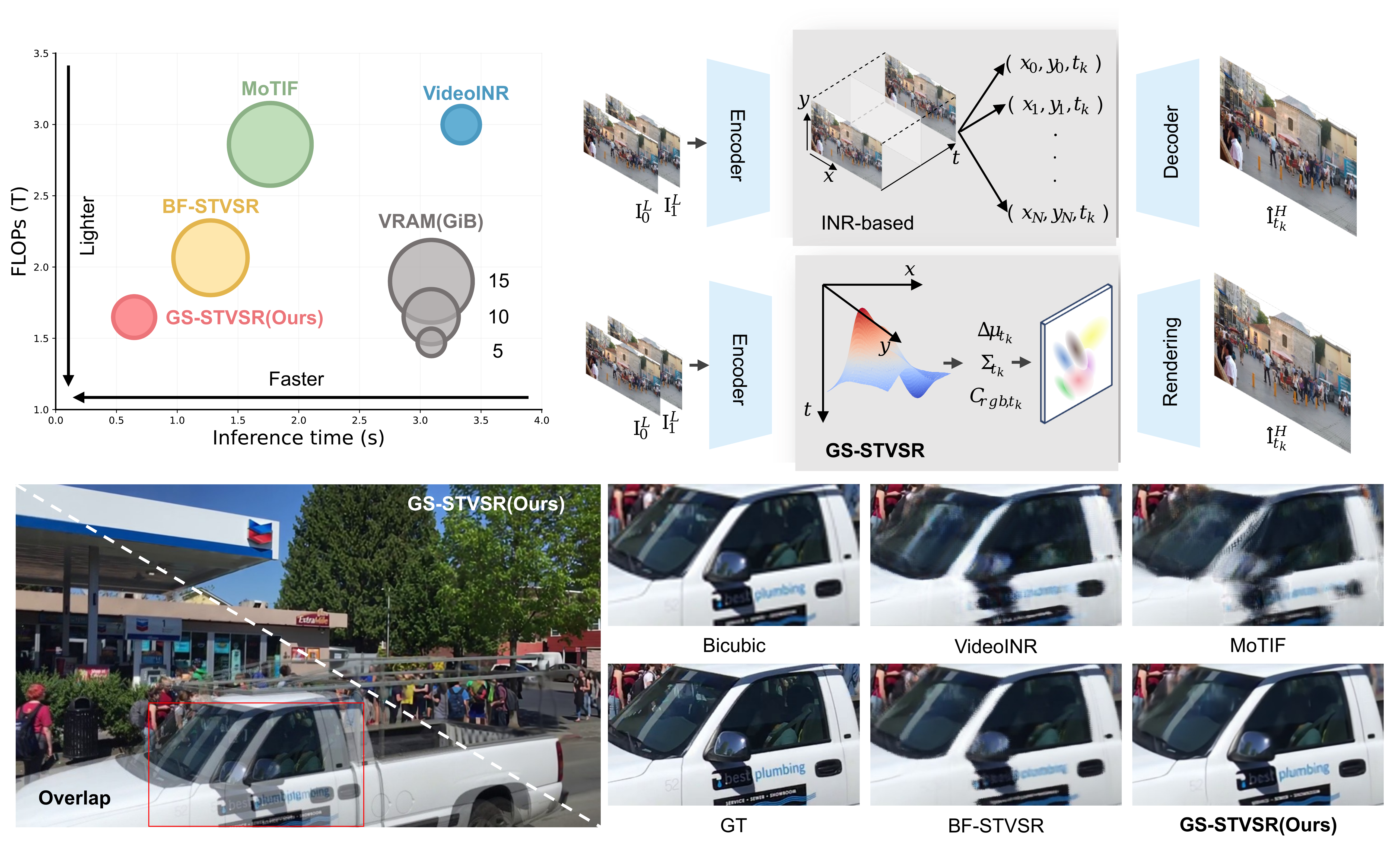}
  \caption{Comprehensive comparison between GS-STVSR and existing C-STVSR methods. \textmd{(Top Left) Comparison of computational efficiency. By comprehensively evaluating computational complexity, inference time, and memory footprint, GS-STVSR significantly outperforms existing methods. (Top Right) Comparison of rendering mechanisms. Previous methods heavily rely on dense pixel-wise queries of INRs. Our method bypasses this bottleneck; once the Gaussian parameters are derived, intermediate frames at arbitrary scales can be rendered with near-zero time cost. (Bottom) Visual comparison among multiple methods under large-scale motion. The figure illustrates the performance of various models in a scenario involving a fast-moving car. Even in such challenging cases, GS-STVSR robustly restores the text on the vehicle. Overall, GS-STVSR maintains extremely high rendering efficiency while delivering highly competitive performance.}}
  \Description{Enjoying the baseball game from the third-base
  seats. Ichiro Suzuki preparing to bat.}
  \label{fig:teaser}
\end{teaserfigure}



\maketitle
\pagestyle{plain}
\thispagestyle{plain}
\begingroup
\renewcommand\thefootnote{}
\footnotetext{$*$Yang Cao and Zhengjun Zha are the corresponding authors. $\dagger$These authors contributed equally to this work. $\ddagger$Xin Di is the project leader.}
\addtocounter{footnote}{-1}
\endgroup

\section{Introduction}

Upgrading low-resolution, low-frame-rate videos to high-fidelity quality is crucial for delivering seamless visual experiences. While traditional Video Super-Resolution (VSR)\cite{caballero2017real, sajjadi2018frame, chan2021basicvsr, chan2022basicvsrpp, rvrt,chen2025dove,zhang2025infvsr,zhuang2025flashvsr,panambur2026creativevr, peng2024lightweight, pengtowards, peng2024efficient, peng2025directing, peng2025boosting, peng2024unveiling, di2025qmambabsr, wu2025hunyuanvideo, fengpmq, wang2023decoupling} and Video Frame Interpolation (VFI)\cite{huang2022rife, niklaus2020softmax, park2021asymmetric, reda2022film, zhang2023extracting,wu2026vtinker,shi2020video,licvpr23amt,0,1,2,3,gong2024beyond,5,6,7} separately address spatial and temporal enhancement, Spatial-Temporal Video Super-Resolution (STVSR)\cite{haris2020space, xiang2020zooming, rstt} aims to accomplish both within a single model. However, most existing STVSR methods operate at fixed scaling factors determined during training (F-STVSR), making them ill-suited for the diverse magnification demands of real-world applications\cite{he2025run, he2025nested, he2024weakly, he2023reti, xiao2024survey, he2023strategic, he2023hqg, he2023camouflaged, he2023degradation, yi2026gdpo, yi2021structure, yi2021efficient, wang2025dual, wang2024ddc, wei2024misalignment, wei2023accurate, wei2025multi, wei2025degradation, wei2025rethinking, xu2025scalar, jin2025semantic, lan2025flux}. To overcome this rigidity, Continuous STVSR (C-STVSR)\cite{chen2022vinr, chen2023motif,kim2025bf} has emerged as a more flexible paradigm that enables arbitrary spatio-temporal upsampling. Realizing such seamless cross-dimensional rendering calls for breaking beyond discrete pixel grids toward truly continuous signal representations.

Driven by this need, Implicit Neural Representation (INR) based methods\cite{liif} have made notable progress in C-STVSR. VideoINR\cite{chen2022vinr} first represented video as a continuous implicit field, learning mappings from spatio-temporal coordinates $(x,y,t)$ to pixel values via MLPs. Subsequent works such as MoTIF\cite{chen2023motif} and BF-STVSR\cite{kim2025bf} further improved quality through better motion modeling and frequency decomposition. Despite these advances, INR methods suffer from two fundamental limitations: (1) dense pixel-wise grid queries whose computational cost grows linearly with the number of interpolated frames, and (2) the spectral bias of MLPs\cite{spectralbias}, which hampers the recovery of high-frequency dynamic details. These bottlenecks severely constrain inference efficiency and hinder practical deployment.

Meanwhile, 2D Gaussian Splatting (2D-GS)\cite{2DGS2024,3Dgaussians,gaussianimage,ptg,GaussianTokenAE,gssrhighfidelity2d} has demonstrated remarkable potential in image super-resolution. PTG\cite{ptg} models each pixel as a 2D Gaussian kernel and leverages the ultra-fast rasterization pipeline to achieve real-time arbitrary-scale super-resolution. Notably, 2D-GS offers two properties that naturally align with C-STVSR: first, its continuous analytic formulation inherently bypasses the dense grid queries required by INR; second, the covariance matrix $\Sigma$ decomposes into scale and rotation components, where the scale directly corresponds to the spatial magnification factor, enabling seamless joint modeling of spatial and temporal dimensions within a unified Gaussian framework.

Building on these insights, we propose GS-STVSR, the first 2D-GS-based framework for C-STVSR. Our framework parameterizes 2D Gaussians as continuous functions of time and drives their spatiotemporal evolution through explicit motion modeling. A key empirical finding underpinning our design is that the covariance parameters of 2D Gaussians exhibit remarkably strong temporal stability (correlation $\sim$0.99 between adjacent frames), far exceeding that of the pixel domain. This allows us to model covariance transitions with ultra-lightweight convolutions while focusing model capacity on the more challenging position and color estimation. Specifically, GS-STVSR introduces three core components: (1) a Covariance Resampling Alignment module that predicts intermediate covariance and re-samples from a pre-defined CPB to prevent covariance drift; (2) an Optical Flow-Guided Continuous Gaussian Motion Learning module that derives Gaussian positions and colors at arbitrary time steps via linear motion interpolation and adaptive feature fusion; and (3) a Motion-Aware Adaptive Offset Window that dynamically adjusts offset ranges based on local motion intensity, enabling accurate trajectory tracking under large displacements.

Our main contributions are summarized as follows:
\begin{itemize}
    \item We propose GS-STVSR, the first framework that extends 2D Gaussian Splatting to C-STVSR, fundamentally bypassing the dense grid query bottleneck of INR-based methods.
    \item We reveal the strong temporal stability of Gaussian covariance parameters and design a Covariance Resampling Alignment module for efficient and stable covariance prediction across time.
    \item We design an optical flow-guided motion learning module coupled with an adaptive offset window to handle continuous Gaussian evolution under diverse motion scales.
    \item Extensive experiments on Vid4, GoPro, and Adobe240 demonstrate that GS-STVSR achieves state-of-the-art performance while delivering over $3\times$ inference speedup at challenging spatiotemporal scales.
\end{itemize}

\begin{figure*}[t!]
    \centering
    \includegraphics[width=\textwidth]{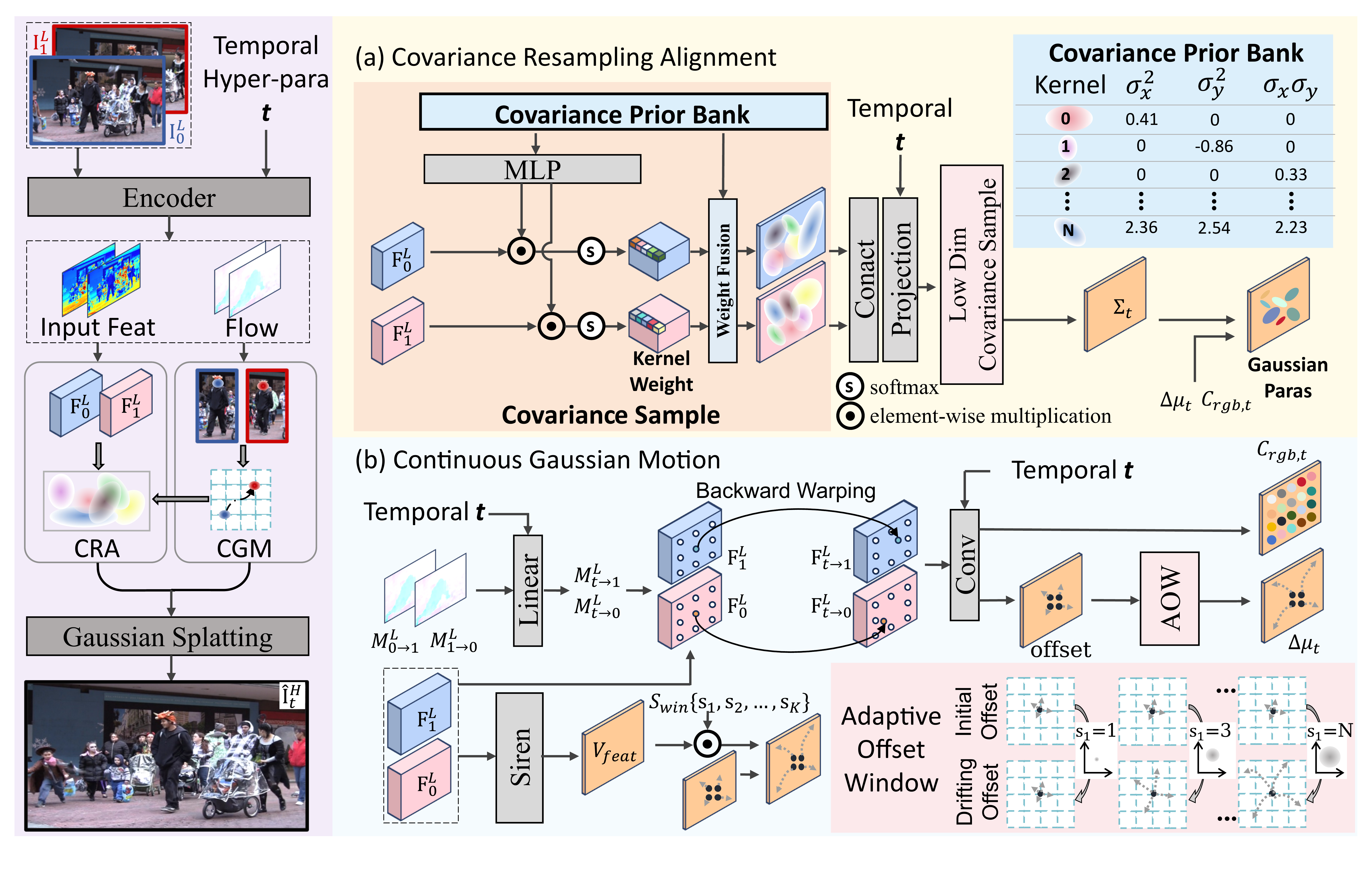}
    \caption{Overview of the GS-STVSR framework.\textmd{Given two input low-resolution frames, we first extract their features and bidirectional optical flows. Based on temporal characteristics, the estimation of 2D Gaussian parameters is decoupled into two branches: (a) The Covariance Resampling Alignment Module predicts the intermediate covariance by resampling from a pre-defined Covariance Prior Bank (CPB), which ensures temporal stability and prevents covariance drift. (b) The Continuous Gaussian Motion Learning Module utilizes intermediate optical flows for feature backward warping to estimate color and position information. Meanwhile, a Motion-Aware Adaptive Offset Window (AOW) dynamically adjusts the position offset range based on local motion intensity to accurately track large-scale motions. Finally, the complete Gaussian parameters are combined with the target spatial scale for fast rendering via 2D Gaussian Splatting rasterization.}}
    \label{fig:GS-STVSR_framework}
\end{figure*}

\section{Related Work}

\subsection{Spatial-Temporal Video Super-Resolution}

Spatial-Temporal Video Super-Resolution (STVSR)\cite{rstt,kim2025bf,chen2023motif,chen2022vinr,xiang2020zooming,xu2021temporal,wei2025evenhancer,lu2024hr,wei2025evenhancerplus} aims to simultaneously enhance the spatial resolution and temporal frame rate of videos. Early efforts typically adopt a two-stage pipeline that cascades Video Frame Interpolation (VFI) models\cite{zheng2025towards, zheng2025decoupled, zheng2024odtrack, zheng2023toward, zheng2022leveraging, duan2025dit4sr, duan2025diffretouch, duan2025diffusion, he2025segment, he2025reversible, xiao2026qualiteacher, Xiao2026Beyond} such as SuperSloMo\cite{superslomo}, DAIN\cite{DAIN}, and QVI\cite{qvi_nips19} with Video Super-Resolution (VSR) models\cite{wan2025attention, Perceive-IR, UniUIR, ClearAIR, zhao2025spike, zhao2022learning, zhao2024optical, li2025ustc, li2024object, bai2025refusion, bai2025task, xiao2025incorporating} such as EDVR\cite{wang2019edvr} and BasicVSR\cite{chan2021basicvsr, chan2022basicvsrpp}. However, such cascaded pipelines suffer from error accumulation between stages and lack joint spatio-temporal optimization, leading to suboptimal results. To address this, one-stage Fixed-scale STVSR (F-STVSR) methods have been proposed to jointly model both dimensions: Haris \textit{et al.} \cite{haris2020space} have introduced a unified framework for addressing STVSR. Xiang \textit{et al.} \cite{xiang2020zooming} have proposed to use bidirectional deformable ConvLSTM. RSTT\cite{rstt} integrates spatial and temporal super-resolution within a unified Transformer, and TMNet\cite{xu2021temporal} further exploits temporal modulation for improved multi-frame reconstruction. Despite their improved performance, these F-STVSR methods are inherently constrained by fixed scaling factors set during training. 
% During inference, the model can only produce outputs at the exact spatial and temporal scales seen during training, making it impractical for real-world scenarios that demand flexible magnification ratios. Moreover, training separate models for each desired scale combination incurs prohibitive computational and storage costs.

To overcome this rigidity, Continuous STVSR (C-STVSR) has emerged to support arbitrary spatio-temporal scaling factors within a single model. VideoINR\cite{chen2022vinr} pioneered this direction by mapping coordinates $(x, y, t)$ to RGB values via MLPs, enabling continuous signal representation. However, its vanilla MLP decoder suffers from spectral bias, limiting its ability to reconstruct high-frequency details. MoTIF\cite{chen2023motif} further replaces backward warping with forward warping using softmax splatting\cite{niklaus2020softmax} and incorporates a pre-trained optical flow network\cite{teed2020raft} to provide explicit motion context, improving temporal alignment but at the cost of additional computational overhead from the auxiliary flow network. BF-STVSR\cite{kim2025bf} decouples spatial and temporal modeling through a B-spline mapper\cite{btc} and a Fourier mapper\cite{tancik2020fourier}, alleviating the spectral bias of vanilla MLPs and achieving state-of-the-art quality among INR-based approaches. However, all these INR-based methods fundamentally rely on dense pixel-wise grid queries, where each output pixel requires an independent forward pass through the MLP decoder. This causes computational overhead to scale linearly with the number of interpolated frames, posing a critical efficiency bottleneck that limits practical deployment. In contrast, our method leverages 2D Gaussian Splatting to drive spatio-temporal Gaussian dynamics, completely bypassing dense grid queries and achieving near-constant inference latency regardless of the temporal scale.

\subsection{Gaussian Splatting}

3D Gaussian Splatting (3D-GS)\cite{3Dgaussians} has emerged as a powerful alternative to NeRF\cite{mildenhall2020nerf} for real-time scene rendering, representing scenes as collections of anisotropic 3D Gaussians that support efficient rasterization and direct scene manipulation. 2D-GS\cite{2DGS2024} further adapts this paradigm to planar representations, improving geometric accuracy through precise 2D Gaussian projection. In the domain of dynamic scene modeling, works such as Dynamic 3D Gaussians\cite{luiten2023dynamic} and 4D Gaussian Splatting\cite{4dgs,TiNeuVox,lei2024mosca,huang2023sc,sgfs,splinegs} track and deform Gaussian primitives over time for dynamic view synthesis, demonstrating the inherent potential of Gaussian representations for temporal modeling, though they primarily target 3D reconstruction and have not been explored for 2D video enhancement tasks. More recently, 2D-GS has been applied to image-level tasks: Zhang \textit{et al.}\cite{gaussianimage} performs image compression and reconstruction via long-term optimization of Gaussian parameters, while Hu \textit{et al.}\cite{gssrhighfidelity2d} operates in the feature space to improve visual quality and speed. However, these methods either require lengthy per-image optimization or still rely on cumbersome upsampling and decoding pipelines\cite{zeng2025eevee, sun2025evdm, xu2024demosaicformer, jiang2024rbsformer, fu2024event, cao2026learning, liu2025event, li2024fouriermamba, zhong2025compevent, liu2025dreamuhd, lu2025evenformer, lu2024mace, lu2023tf, lu2024robust, lu2025does, zhou2025dragflow, li2025set, gao2024eraseanything, lu2022copy, ren2025all, yang2025temporal, gao2025revoking, yu2025visual, zhao2026luve, zhu2024oftsr, 11316813, li2023cross, li2024efficient, li2024efficientSR, li2025dual, li2025self, li2025survey, li2026seeing, li2026fouriersr, li2025measurement, gao2022feature}, struggling to reconstruct continuous high-resolution signals efficiently. Peng \textit{et al.}\cite{ptg} breaks this limitation by revealing the Covariance Prior Bank(CPB)---the finding that covariance parameters of natural images follow a low-dimensional smooth distribution---and proposes CPB-Driven Covariance Weighting to achieve ultra-fast arbitrary-scale image super-resolution via efficient 2D-GS rasterization, completely discarding traditional upsampling decoders. Inspired by PTG, our work extends 2D-GS to the substantially more challenging C-STVSR task, where the key difficulty lies in modeling how each Gaussian kernel's position, color, and covariance evolve continuously over time while preserving the efficiency of Gaussian rasterization.

\section{Method}

\subsection{Overview}
The overall pipeline of GS-STVSR is illustrated in Fig.~\ref{fig:GS-STVSR_framework}. Given two low-resolution frames $I_0^L, I_1^L \in \mathbb{R}^{3 \times H \times W}$, our goal is to generate a high-resolution intermediate frame $I_t^H \in \mathbb{R}^{3 \times sH \times sW}$ at arbitrary time $t \in [0, 1]$ with arbitrary spatial scale $s$. Unlike INR-based methods that query a dense coordinate grid to reconstruct each output pixel, we adopt 2D Gaussian Splatting as our rendering primitive, representing each frame as a collection of 2D Gaussian kernels whose parameters evolve continuously over time. Following PTG\cite{ptg}, we reconstruct the continuous signal $f_c(x, y, t)$ as a superposition of $N$ Gaussian kernels:
\begin{equation}
f_c(x, y, t) = \sum_{i=1}^N G_i(x, y, t).
\end{equation}
Each Gaussian kernel $G_i$ is parameterized by eight degrees of freedom: a $2 \times 2$ covariance matrix $\Sigma_t$ that encodes the spatial extent and orientation, a 2D position vector $\mu_t$ indicating the kernel center, and a 3D RGB color vector $C_{rgb,t}$:
\begin{equation}
\Sigma_t = \begin{bmatrix} \sigma_{x,t}^2 & \rho_t \sigma_{x,t} \sigma_{y,t} \\ \rho_t \sigma_{x,t} \sigma_{y,t} & \sigma_{y,t}^2 \end{bmatrix}, \quad \mu_t = \begin{bmatrix} \mu_{x,t} \\ \mu_{y,t} \end{bmatrix}, \quad C_{rgb,t} = \begin{bmatrix} C_{r,t} \\ C_{g,t} \\ C_{b,t} \end{bmatrix}.
\end{equation}
The value of $G_i$ at position $(x, y)$ and time $t$ is given by:
\begin{equation}
G_i(x, y, t) = C_{rgb,t} \cdot \frac{1}{2\pi |\Sigma_{i,t}|} \exp\!\left(-\frac{1}{2} d_t^\top \Sigma_{i,t}^{-1} d_t\right),
\end{equation}
where $d_t = (x, y)^\top - \mu_t$ is the displacement from the kernel center. A key advantage of this formulation is that the covariance matrix $\Sigma_t$ can be decomposed into scale and rotation components, where the scale directly corresponds to the spatial magnification factor. This means that once the Gaussian parameters are predicted at the LR resolution, rendering at any HR scale $s$ is achieved simply by adjusting the scale component of $\Sigma_t$ during rasterization, without re-running the network. Following PTG, the position $\mu$ is fixed at the LR pixel center and a learnable offset $\Delta \mu$ is predicted via Adaptive Position Drifting. Our goal thus reduces to estimating $\{\Delta \mu_t, \Sigma_t, C_{rgb,t}\}$ for arbitrary $t \in [0,1]$.

We employ an encoder $E$ to extract latent features $F_0^L, F_1^L \in \mathbb{R}^{C \times H \times W}$ from the input LR frames, and a pre-trained optical flow extractor to obtain bidirectional motion vectors $M_{0 \to 1}^L, M_{1 \to 0}^L \in \mathbb{R}^{2 \times H \times W}$. The Gaussian parameter estimation is then decoupled into two branches based on distinct temporal characteristics: (1) the covariance $\Sigma_t$, which exhibits strong temporal stability, is derived from endpoint covariances $\Sigma_0, \Sigma_1$ via a lightweight CPB re-sampling alignment module (Sec.~3.2); (2) the position offset $\Delta \mu_t$ and color $C_{rgb,t}$, which are more sensitive to motion, are generated through an optical flow-guided motion learning module (Sec.~3.3). An adaptive offset window (Sec.~3.4) further adjusts $\Delta \mu_t$ to handle large-scale motion. Finally, the complete Gaussian parameters are combined with the target spatial scale $s$ for fast 2D-GS rasterization.

% \begin{figure}[htbp]
%     \centering
%     \includegraphics[width=0.35\textwidth]{measure_index.pdf}
%     \caption{Temporal correlation comparison between pixel-domain values and Gaussian covariance parameters.\textmd{Red: covariance parameters; Blue: pixel domain. Solid lines: Pearson correlation coefficient; Dashed lines: cosine similarity.}}
%     \Description{Comparisons of Temporal correlation between the pixel domain and covariance parameters.Red represents the covariance parameters, blue represents the pixel domain, and the solid lines measure the Pearson correlation coefficient, while the dashed lines measure the cosine similarity.}
%     \label{fig:measure_index}
% \end{figure}

\subsection{Covariance Resampling Alignment Module}

% We obtain the endpoint covariance parameters $\Sigma_0$ and $\Sigma_1$ by mapping the deep features $F_0^L$ and $F_1^L$ to sample from a pre-defined Covariance Prior Bank (CPB), following the methodology in \cite{ptg}. The Covariance prior reveals that the covariance parameters of natural images follow a traceable, low-dimensional smooth probability space distribution. Building on this prior, a key observation motivating our design is the remarkably strong temporal stability of covariance parameters. We conduct experiments to measure the temporal correlation of covariance parameters and pixel-domain values at varying time intervals ($\Delta t \in \{1, 3, 5, 7\}$) using both Pearson Correlation Coefficient and Cosine Similarity. As shown in Fig.~\ref{fig:measure_index}, as the time interval increases from 1 to 7 frames, the pixel-domain cosine similarity drops sharply, while covariance parameters remain remarkably stable around 0.99 across both metrics. We attribute this stability to the fact that the macroscopic geometric structure captured by 2D Gaussian covariance is inherently dominated by low-frequency information, whose evolution along the time axis is smooth and predictable. This important insight implies that characterizing the temporal transition of covariance does not require complex non-linear transformations; lightweight operations are theoretically sufficient.

We obtain the endpoint covariance parameters $\Sigma_0$ and $\Sigma_1$ by mapping the deep features $F_0^L$ and $F_1^L$ to sample from a pre-defined Covariance Prior Bank (CPB), following the methodology in \cite{ptg}. The Covariance prior reveals that the covariance parameters of natural images follow a traceable, low-dimensional smooth probability space distribution. Building on this prior, a key observation motivating our design is the remarkably strong temporal stability of covariance parameters. We conduct experiments to measure the temporal correlation of covariance parameters and pixel-domain values at varying time intervals using both Pearson Correlation Coefficient and Cosine Similarity. As the time interval increases, the pixel-domain similarity drops sharply, while the covariance parameters remain remarkably stable across both metrics. Detailed experimental data and analysis are provided in the supplementary material. We attribute this stability to the fact that the macroscopic geometric structure captured by 2D Gaussian covariance is inherently dominated by low-frequency information, whose evolution along the time axis is smooth and predictable. This important insight implies that characterizing the temporal transition of covariance does not require complex non-linear transformations; lightweight operations are theoretically sufficient.

Based on this insight, we predict the intermediate covariance $\Sigma_t$ via a single-layer convolution on the concatenation of $\Sigma_0$, $\Sigma_1$, and the target time $t$ along the channel dimension. However, directly outputting high-dimensional covariance parameters may cause covariance drift, where unconstrained convolution operations perturb the original covariance features, causing the predicted $\Sigma_t$ to deviate from the low-dimensional smooth manifold strictly defined by the CPB $\mathcal{K}$. To strike a balance between computational efficiency and optimization stability, we instead compress the convolution output into a low-dimensional fusion feature $E_t$:
\begin{equation}
E_t = \mathrm{Conv}(\mathrm{Concat}[\Sigma_0, \Sigma_1, t]),
\end{equation}
and treat $E_t$ as adaptive sampling weights over the CPB. Following PTG\cite{ptg}, the intermediate covariance is obtained via weighted re-combination:
\begin{equation}
\Sigma_t = \sum_{i=1}^{N} w_{i,t}\, G_i, \quad W_t = \mathrm{Softmax}(E_t),
\end{equation}
where $G_i \in \mathcal{K}$ is a predefined Gaussian kernel and $W_t$ the corresponding weight. This re-sampling operation anchors $\Sigma_t$ within the CPB manifold, eliminating covariance drift while preserving ultra-low computational overhead.

\subsection{Optical Flow-Guided Continuous Gaussian Motion Learning}

To generate the position and color parameters $\Delta \mu_t, C_{rgb,t}$ at arbitrary time $t$, we design an optical flow-guided motion learning module that replaces the MLP-based parameter generation in PTG\cite{ptg}. Directly reusing PTG's MLP for temporal extension would lack explicit motion awareness and require completely passing through the massive MLP network for every new time step, contradicting our efficiency goal. Instead, we leverage pre-extracted bidirectional optical flows $M_{0 \to 1}^L, M_{1 \to 0}^L \in \mathbb{R}^{2 \times H \times W}$ to provide explicit motion guidance. Under the linear motion assumption within short inter-frame intervals, the intermediate optical flows pointing from time $t$ to the endpoint frames are obtained by simple scaling:
\begin{equation}
M_{t \to 1}^L = t \cdot M_{1 \to 0}^L, \quad M_{t \to 0}^L = (1 - t) \cdot M_{0 \to 1}^L.
\end{equation}
With the estimated intermediate flows, the endpoint deep features $F_0^L, F_1^L$ are temporally aligned to the target moment $t$ via backward warping:
\begin{equation}
F_{0,t}^L = \mathcal{W}(F_0^L,\, M_{t \to 0}^L), \quad F_{1,t}^L = \mathcal{W}(F_1^L,\, M_{t \to 1}^L),
\end{equation}
where $\mathcal{W}(\cdot,\cdot)$ denotes the backward warping operator. Since the linear motion assumption may introduce minor alignment errors, particularly around occlusion boundaries, we feed the warped features into a lightweight convolution module that jointly predicts an adaptive spatio-temporal fusion mask $M \in [0,1]$ and a feature residual $\Delta F$ to compensate for these inaccuracies\cite{res}. The fused intermediate feature is computed as:
\begin{equation}
F_t^L = M \odot F_{0,t}^L + (1 - M) \odot F_{1,t}^L + \Delta F,
\end{equation}
where $\odot$ denotes element-wise multiplication. The mask $M$ adaptively balances the contribution of the two warped features at each spatial location, while $\Delta F$ captures residual details that cannot be recovered by warping alone. Finally, the fused feature $F_t^L$ is decoded by a lightweight convolutional head to produce the Gaussian parameters $\Delta \mu_t$ and $C_{rgb,t}$. This design avoids the heavy per-frame MLP inference required by INR methods, as the optical flow computation is shared across all time steps and only the lightweight fusion and decoding stages are time-dependent.

\begin{figure}[t]
    \centering
    % --- 第一个子图 ---
    \begin{subfigure}[t]{0.15\textwidth}
        \centering
        \includegraphics[width=\textwidth]{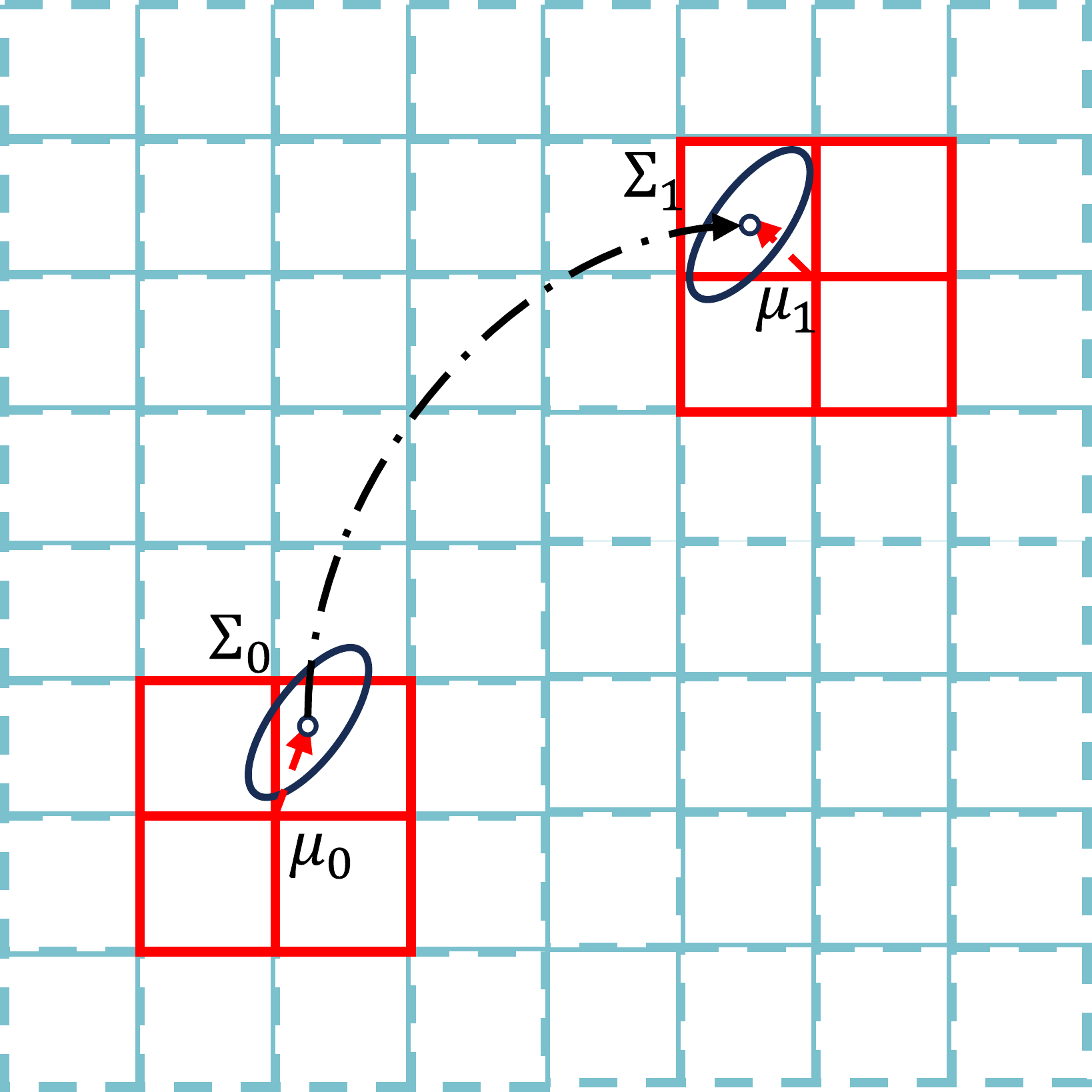} % 替换为你的图片名
        \caption{Initial and final states}
        \label{fig:sub1}
    \end{subfigure}
    \hfill % \hfill 用于在两个子图之间均匀撑开空白
    % --- 第二个子图 ---
    \begin{subfigure}[t]{0.15\textwidth}
        \centering
        \includegraphics[width=\textwidth]{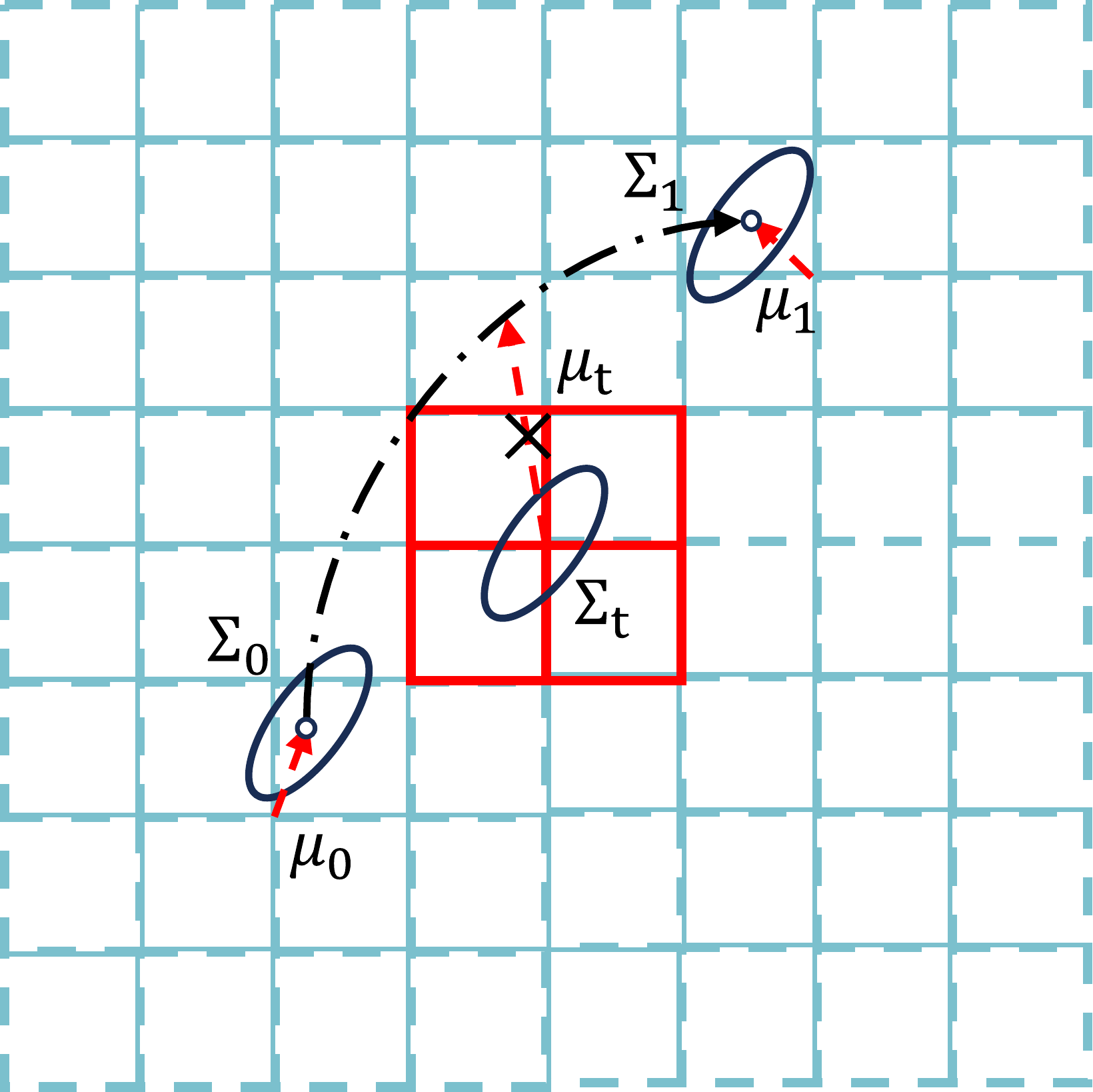} % 替换为你的图片名
        \caption{Interpolated state with small offset}
        \label{fig:sub2}
    \end{subfigure}
    \hfill % \hfill 用于在两个子图之间均匀撑开空白
    \begin{subfigure}[t]{0.15\textwidth}
        \centering
        \includegraphics[width=\textwidth]{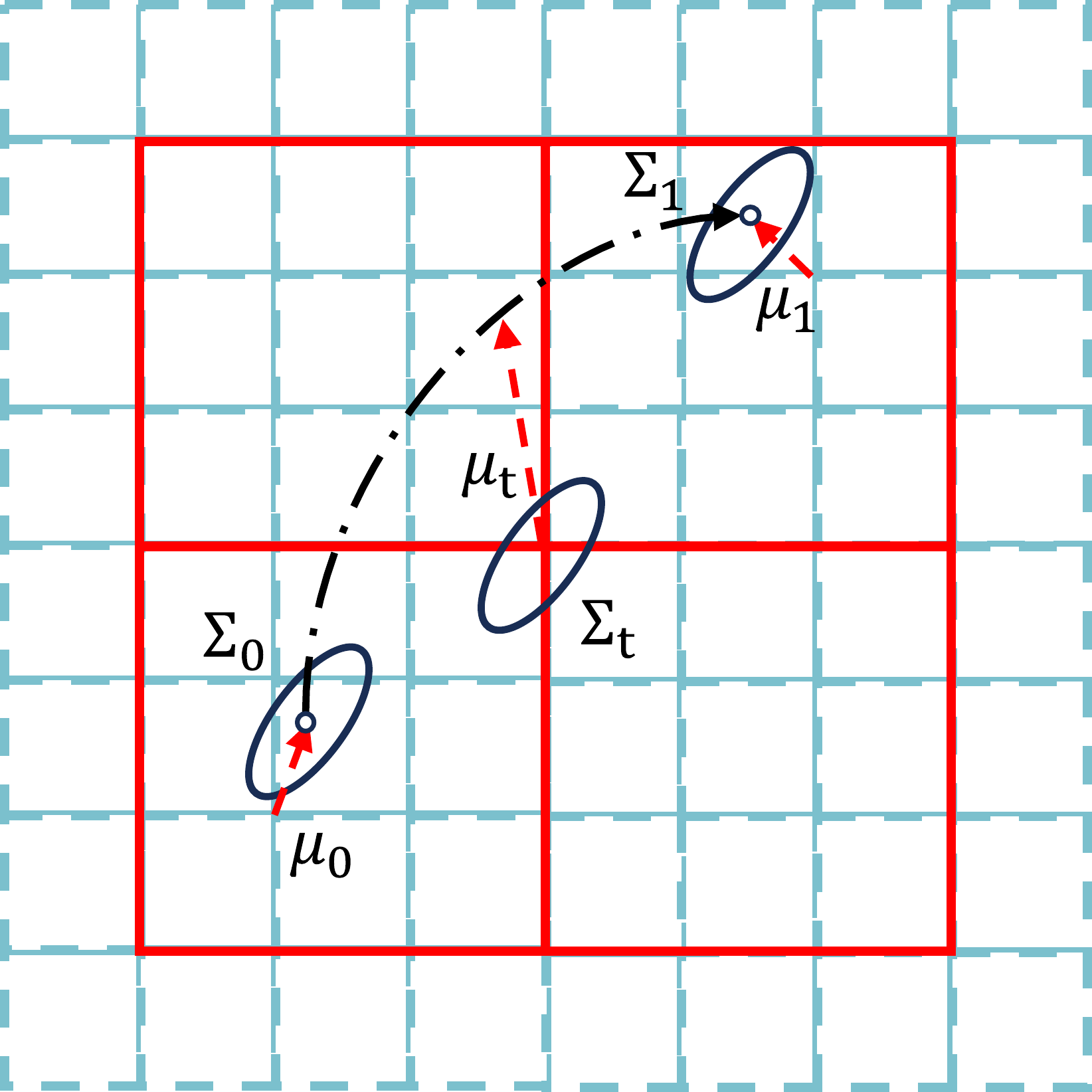} % 替换为你的图片名
        \caption{Interpolated state with large offset}
        \label{fig:sub3}
    \end{subfigure}
    % --- 总图例 ---
    \caption{Illustration of the Motion-Aware Adaptive Offset Window under large-scale motion.\textmd{(\subref{fig:sub1}) Possible trajectories between endpoint frames. (\subref{fig:sub2}) A small offset window provides insufficient range to reach the actual trajectory. (\subref{fig:sub3}) An appropriately large offset window ensures accurate trajectory targeting.}}
    \label{fig:motion_overall}
\end{figure}

\subsection{Motion-Aware Adaptive Offset Window}

PTG\cite{ptg} restricts the position offset to $\Delta\mu \in [0,1]$, confining each Gaussian kernel to its local pixel grid for training stability. While effective for static images, this constraint creates a critical bottleneck when extended to video: large inter-frame motion requires Gaussian kernels to shift substantially along their trajectories, far exceeding the single-pixel range. With the introduction of the temporal axis, the positional offset should inherently carry a strong temporal motion correlation and be responsible for guiding the continuous movement of the Gaussian kernels. When the offset window is strictly constrained to a minimal range, this hard truncation severely disrupts the temporal correlation of the offsets, making it difficult for the model to guide the Gaussian kernels in tracking large-scale local motion. As illustrated in Fig.~\ref{fig:motion_overall}, when the offset window is too narrow, the model cannot guide Gaussian kernels to track objects undergoing large displacements, resulting in severe multi-frame overlapping artifacts. A naive global relaxation of the window, however, does not solve the problem---our preliminary experiments show that uniformly enlarging the offset range actually hurts performance by excessively expanding the Gaussian degrees of freedom in static regions, leading to unstable optimization.

We observe that the required offset range is inherently tied to local motion intensity: regions with significant motion demand wider windows, while static or slowly-moving regions benefit from tight constraints. This motion prior can be readily estimated from the input endpoint frames. Accordingly, we propose a Motion-Aware Adaptive Offset Window that spatially varies the offset range. Specifically, we learn a weight map $V_{map}$ from the endpoint features $F_0^L, F_1^L$ through Sirens \cite{siren} and combine it with a set of predefined base windows $S_{win} = \{s_1, s_2, \dots, s_K\}$ to produce a spatially-varying window map:
\begin{equation}
W_{map} = \sum_{i=1}^{K} V_{map,i} \cdot s_i,
\end{equation}
where $S_{win} = \{1, 2, \dots, 10\}$ in our experiments. Note that $W_{map}$ is computed once per input pair and shared across all intermediate time steps, introducing negligible overhead. The final offset is obtained by scaling the initial prediction:
\begin{equation}
\Delta\mu_t \leftarrow \Delta\mu_t \odot W_{map},
\end{equation}
where $\odot$ denotes element-wise multiplication. This mechanism adaptively grants each spatial location an offset range proportional to its local motion magnitude, enabling accurate trajectory tracking for large-scale motion while maintaining tight regularization in static regions. The complete set of Gaussian parameters $\{\Delta\mu_t, \Sigma_t, C_{rgb,t}\}$ at arbitrary time $t$ is then combined with the target spatial scale $s$ for fast rendering via 2D-GS rasterization.

\section{Experiments}

\subsection{Experiments Setup}

\begin{table*}[htbp]
\centering
\caption{Performance comparison on the F-STVSR baselines on Vid4, GoPro\cite{Nah_2017_CVPR}, and Adobe240\cite{adobe240} datasets.\textmd{Results are evaluated using PSNR (dB) and SSIM metrics in Y channels. All frames are interpolated by a factor of $\times 4$ in the spatial axis and $\times 8$ in the temporal axis. ``Average'' refers to metrics calculated across all 8 interpolated frames, while ``Center'' refers to metrics measured using $1^{st}$, $4^{th}$ and $9^{th}$ (that is the center-frame interpolation) frames of the interpolated sequence. "AT" represents the average time consumed for a single inference on Adobe240. \textcolor{red}{Red} and \textcolor{blue}{blue} indicate the best and the second best performance, respectively.}}
\label{tab:t1} 
\resizebox{\textwidth}{!}{
\begin{tabular}{ccccccccc}
\toprule
\shortstack{VFI Method} & VSR Method & Vid4 & GoPro-Center & GoPro-Average & Adobe-Center & Adobe-Average & Parameters (M) & AT (s)\\
\midrule
SuperSloMo\cite{superslomo} & Bicubic & 22.42 / 0.5645 & 27.04 / 0.7937 & 26.06 / 0.7720 & 26.09 / 0.7435 & 25.29 / 0.7279 & 19.8 & -\\
SuperSloMo\cite{superslomo}& EDVR\cite{wang2019edvr} & 23.01 / 0.6136 & 28.24 / 0.8322 & 26.30 / 0.7960 & 27.25 / 0.7972 & 25.90 / 0.7682 & 19.8+20.7 & -\\
SuperSloMo\cite{superslomo} & BasicVSR\cite{chan2021basicvsr} & 23.17 / 0.6159 & 28.23 / 0.8308 & 26.36 / 0.7977 & 27.28 / 0.7961 & 25.94 / 0.7679 & 19.8+6.3 & - \\
\addlinespace
QVI\cite{qvi_nips19} & Bicubic & 22.11 / 0.5498 & 26.50 / 0.7791 & 25.41 / 0.7554 & 25.57 / 0.7324 & 24.72 / 0.7114 & 29.2 & - \\
QVI\cite{qvi_nips19} & EDVR\cite{wang2019edvr} & 23.48 / 0.6547 & 28.60 / 0.8417 & 26.64 / 0.7977 & 27.45 / 0.8087 & 25.64 / 0.7590 & 29.2+20.7 & - \\
QVI\cite{qvi_nips19} & BasicVSR\cite{chan2021basicvsr} & 23.15 / 0.6428 & 28.55 / 0.8400 & 26.27 / 0.7955 & 26.43 / 0.7682 & 25.20 / 0.7421 & 29.2+6.3 & - \\
\addlinespace
DAIN\cite{DAIN} & Bicubic & 22.57 / 0.5732 & 26.92 / 0.7911 & 26.11 / 0.7740 & 26.01 / 0.7461 & 25.40 / 0.7321 & 24.0 & - \\
DAIN\cite{DAIN} & EDVR\cite{wang2019edvr} & 23.48 / 0.6547 & 28.58 / 0.8417 & 26.64 / 0.7977 & 27.45 / 0.8087 & 25.64 / 0.7590 & 24.0+20.7 & - \\
DAIN\cite{DAIN} & BasicVSR\cite{chan2021basicvsr} & 23.43 / 0.6514 & 28.46 / 0.7966 & 26.43 / 0.7966 & 26.23 / 0.7725 & 25.23 / 0.7725 & 24.0+6.3 & - \\
\addlinespace
\multicolumn{2}{c}{ZoomingSloMo\cite{xiang2020zooming}} & 25.72 / 0.7717 & 30.69 / 0.8847 & - / - & 30.26 / 0.8821 & - / - & 11.10 & - \\
\multicolumn{2}{c}{TMNet\cite{xu2021temporal}} & \textcolor{blue}{25.96 / 0.7803} & 30.14 / 0.8696 & 28.83 / 0.8514 & 29.41 / 0.8524 & 28.30 / 0.8354 & 12.26 & 5.67 \\
\addlinespace
\multicolumn{2}{c}{VideoINR\cite{chen2022vinr}} & 25.61 / 0.7709 & 30.26 / 0.8792 & 29.41 / 0.8669 & 29.92 / 0.8746 & 29.27 / 0.8651 & 11.31 & 3.34 \\
\multicolumn{2}{c}{MoTIF\cite{chen2023motif}} & 25.79 / 0.7745 & 31.04 / 0.8877 & 30.04 / 0.8773 & 30.63 / 0.8839 & 29.82 / 0.8750 & 12.55 & 1.77 \\
\addlinespace
\multicolumn{2}{c}{BF-STVSR\cite{kim2025bf}} & 25.85 / 0.7772 & \textcolor{blue}{31.17 / 0.8898} & \textcolor{blue}{30.22 / 0.8802} & \textcolor{blue}{30.83 / 0.8880} & \textcolor{blue}{30.12 / 0.8808} & 13.47 & \textcolor{blue}{1.27}\\
\multicolumn{2}{c}{Ours} & \textcolor{red}{26.04 / 0.7822} & \textcolor{red}{31.33 / 0.8918} & \textcolor{red}{30.35 / 0.8817} & \textcolor{red}{31.13 / 0.8907} & \textcolor{red}{30.35 / 0.8827} & {12.67} & \textcolor{red}{0.64}\\
\bottomrule
\end{tabular}
}
\end{table*}

\begin{table*}[htbp]
\centering
\caption{Performance comparison on the C-STVSR baselines for out-of-distribution scale on GoPro\cite{Nah_2017_CVPR} dataset.\textmd{Results are evaluated using PSNR (dB) and SSIM metrics in Y channels. All frames are interpolated by a scaling factor specified on the table and metrics calculated across all interpolated frames. \textcolor{red}{Red} and \textcolor{blue}{blue} indicate the best and the second best performance, respectively.}}
\label{tab:t2} 
\resizebox{\textwidth}{!}{
\begin{tabular}{cccccccccc}
\toprule
% 第一层表头: 处理跨两行的单元格, 以及跨两列的主标题
Temporal & Spatial & \multicolumn{2}{c}{RIFE\cite{huang2022rife}} & \multicolumn{2}{c}{EMA-VFI\cite{zhang2023extracting}} & \multirow{2}{*}{VideoINR\cite{chen2022vinr}} & \multirow{2}{*}{MoTIF\cite{chen2023motif}} & \multirow{2}{*}{BF-STVSR\cite{kim2025bf}} & \multirow{2}{*}{Ours} \\
Scale & Scale & LIIF\cite{liif} & LTE\cite{lte} & LIIF\cite{liif} & LTE\cite{lte} & & & &\\
\midrule
$\times 8$ & $\times 4$ & 29.14 / 0.8524& 29.14 / 0.8524 & 29.68 / 0.8671 & 29.68 / 0.8667 & 29.41 / 0.8669 & 30.04 / 0.8773 & \textcolor{blue}{30.22 / 0.8802} & \textcolor{red}{30.33 / 0.8815} \\
\addlinespace
\multirow{3}{*}{$\times 6$} & $\times 4$ & 30.16 / 0.8738 & 30.16 / 0.8737 & 30.64 / 0.8850 & 30.64 / 0.8848 & 30.75 / 0.8978 & 31.53 / 0.9060 & \textcolor{blue}{31.71 / 0.9082} & \textcolor{red}{31.83 / 0.9092} \\
& $\times 6$ & 27.87 / 0.8038 & 27.86 / 0.8031 & 28.17 / 0.8126 & 28.17 / 0.8117 & 28.37 / 0.8368 & 29.25 / 0.8499 & \textcolor{blue}{29.33 / 0.8514} & \textcolor{red}{29.39 / 0.8517} \\
& $\times 12$ & 24.74 / 0.7019 & 24.70 / 0.6994 & 24.85 / 0.7052 & 24.82 / 0.7028 & 24.55 / 0.7126 & \textcolor{blue}{25.62 / 0.7310} & 25.56 / 0.7276 & \textcolor{red}{25.74 / 0.7341} \\
\addlinespace
\multirow{3}{*}{$\times 12$} & $\times 4$ & 27.43 / 0.8102 & 27.42 / 0.8100 & 27.90 / 0.8263 & 27.90 / 0.8260 & 27.36 / 0.8174 & 27.73 / 0.8222 & \textcolor{blue}{28.08 / 0.8287} & \textcolor{red}{28.18 / 0.8303} \\
& $\times 6$ & 26.19 / 0.7640 & 26.19 / 0.7636 & 26.49 / 0.7748 & 26.49 / 0.7743 & 26.14 / 0.7795 & 26.68 / 0.7897 & \textcolor{blue}{26.96 / 0.7955} & \textcolor{red}{27.10 / 0.7979} \\
& $\times 12$ & 24.03 / 0.6869 & 24.00 / 0.6853 & 24.16 / 0.6918 & 24.15 / 0.6902 & 23.58 / 0.6902 & 24.56 / \textcolor{blue}{0.7088} & \textcolor{blue}{24.68} / 0.7084 & \textcolor{red}{24.88 / 0.7155} \\
\addlinespace
\multirow{3}{*}{$\times 16$} & $\times 4$ & 26.08 / 0.7735 & 26.08 / 0.7733 & \textcolor{red}{26.56 / 0.7904} & \textcolor{blue}{26.56 / 0.7902} & 25.76 / 0.7738 & 25.93 / 0.7745 & 26.40 / 0.7839 &  26.53 / 0.7866  \\
& $\times 6$ & 25.24 / 0.7394 & 25.24 / 0.7391 & 25.54 / 0.7503 & 25.55 / 0.7499 & 25.01 / 0.7473 & 25.24 / 0.7513 & \textcolor{blue}{25.71 / 0.7611} & \textcolor{red}{25.85 / 0.7642} \\
& $\times 12$ & 23.57 / 0.6781 & 23.56 / 0.6769 & 23.68 / 0.6828 & 23.69 / 0.6816 & 23.00 / 0.6762 & 23.73 / 0.6903 & \textcolor{blue}{24.04 / 0.6936} & \textcolor{red}{24.23 / 0.7009} \\
\bottomrule
\end{tabular}
}
\end{table*}

\begin{figure*}[htbp]
    \centering
    % --- 第一个子图 ---
    \begin{subfigure}[t]{0.45\textwidth}
        \centering
        \includegraphics[width=\textwidth]{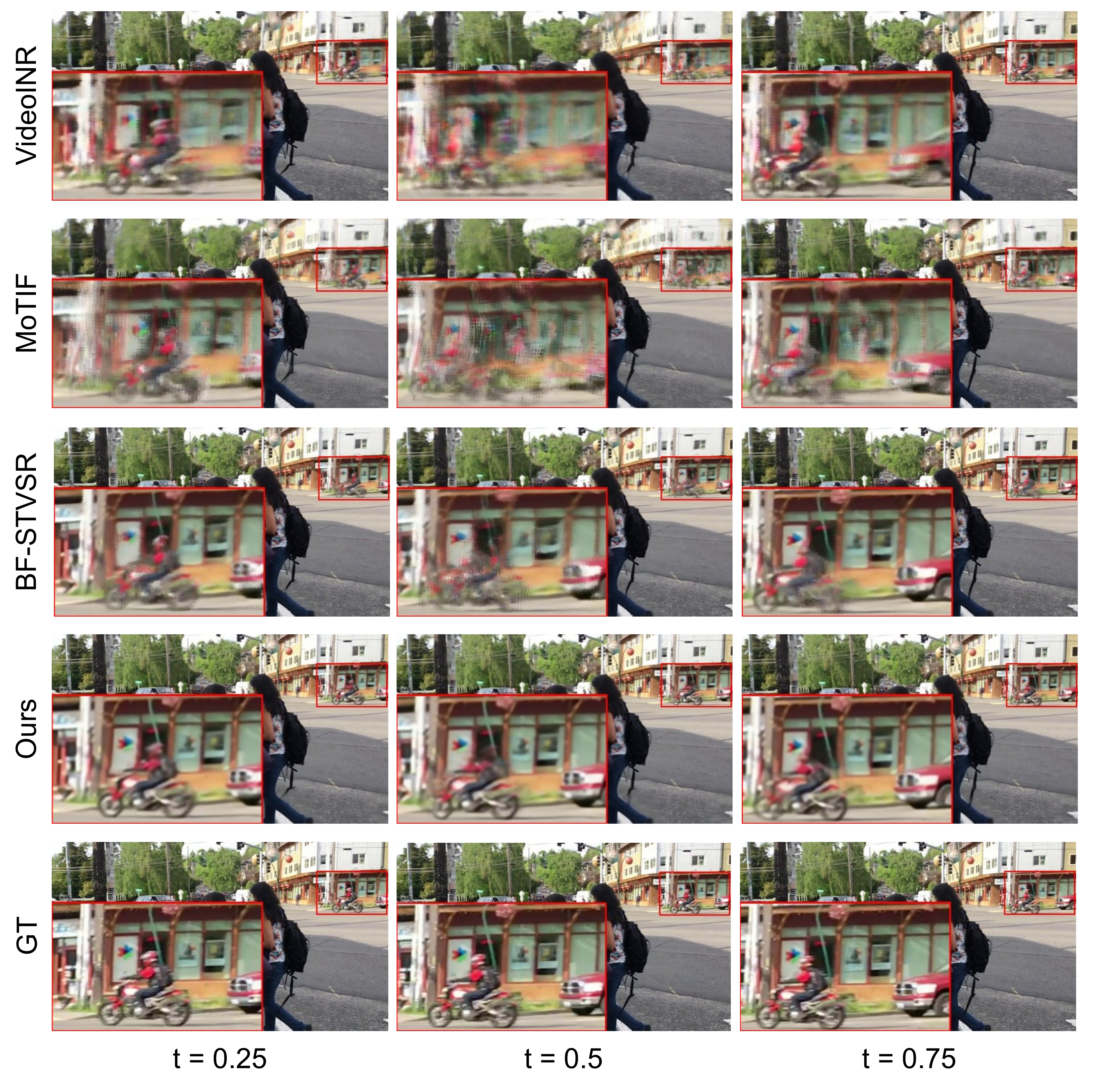}
        \caption{In Distribution}
        \label{fig:sub1}
    \end{subfigure}
    % \hfill % \hfill 用于在两个子图之间均匀撑开空白
    % --- 第二个子图 ---
    \begin{subfigure}[t]{0.45\textwidth}
        \centering
        \includegraphics[width=\textwidth]{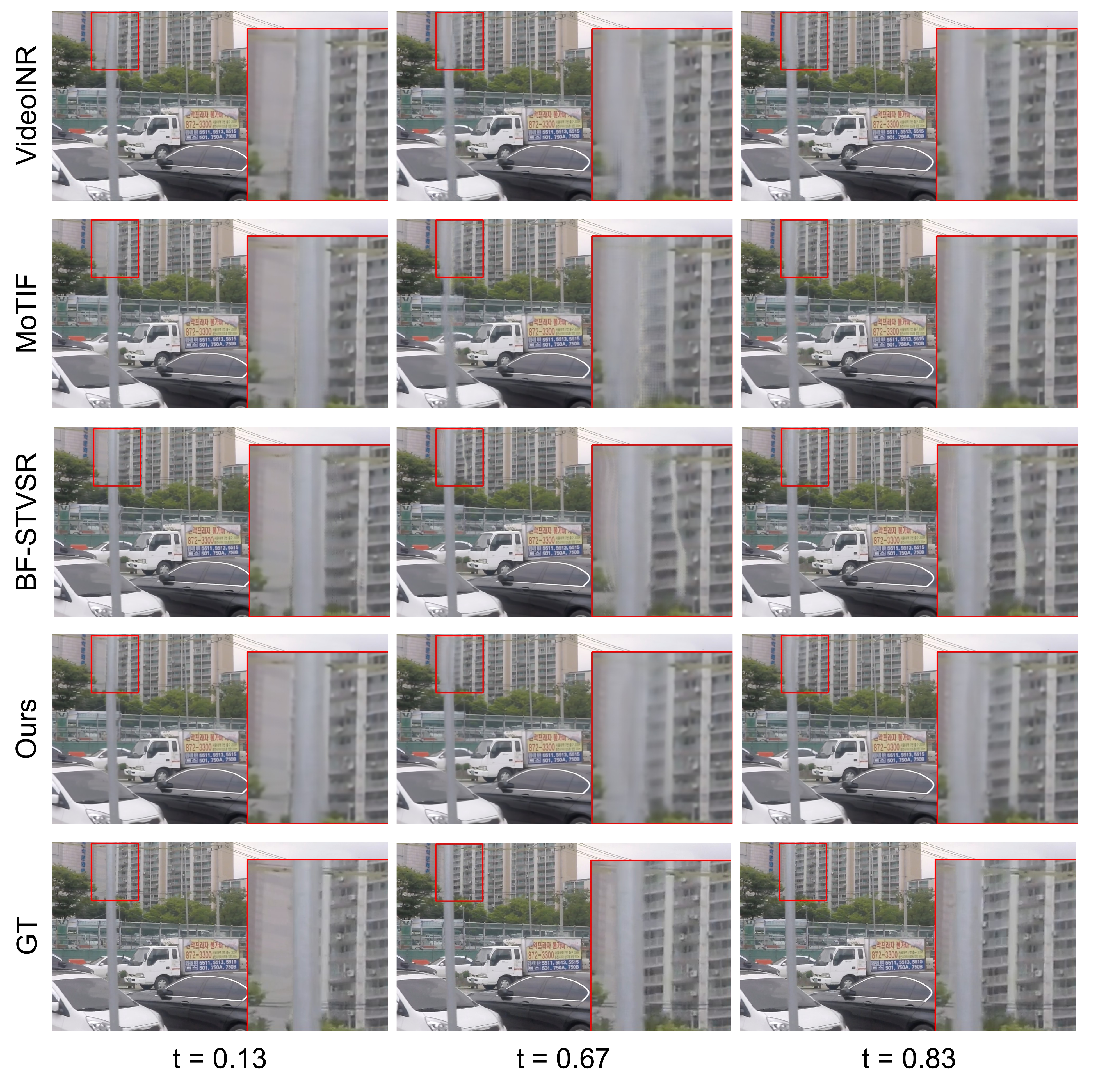}
        \caption{Out of Distribution}
        \label{fig:sub2}
    \end{subfigure}
    % --- 总图例 ---
    \caption{Qualitative comparison on arbitrary-scale temporal interpolation.\textmd{ (\subref{fig:sub1}) shows the visualization results on in-distribution temporal scale ($\times 8$), used during training. (\subref{fig:sub2}) shows the visualization results on out-of-distribution temporal scale ($\times 6$), not used during training.}}
    \Description{Within the training distribution, we presented a scene of a fast-moving motorcycle. Our model was able to smoothly interpolate the position of the motorcycle rider at any intermediate frame, even if there was local deformation, but most of the information was retained. However, the interpolation effect of the C-STVSR model in this scenario was nearly ruined. Outside the training distribution, we presented a scene where objects moved in front of each other and blocked each other. Our model remained stable and did not distort the foreground and background information due to object occlusion. In contrast, the other models all showed distortion phenomena, and the overlapping areas had obvious pixelated patterns, affecting the visual perception.}
    \label{fig:visual}
\end{figure*}

\subsubsection{Implementation and Training Details.} 
We adopt a single-stage training strategy, uniformly sampling the spatial scale from $[4, 8]$ and fixing the temporal scale to $\times 8$ with 3 randomly sampled intermediate time steps. The model is trained for 600K iterations on 8 NVIDIA Tesla V100 GPUs using the Adam\cite{2014Adam} optimizer ($\beta_1 = 0.9$, $\beta_2 = 0.999$) with cosine annealing that decays the learning rate from $10^{-4}$ to $10^{-7}$ every 100K iterations; the learning rate is further scaled by $0.75$ and $0.5$ for the last 200K iterations. We employ RVRT\cite{rvrt} as the feature encoder and reuse its internal SpyNet\cite{spynet2017} as the optical flow extractor to minimize additional parameters. The batch size is 32 with a patch size of $256 \times 256$, and standard data augmentation (random rotation and horizontal flipping) is applied. The training loss combines L1 loss\cite{liif} and frequency loss\cite{cui2023focal} with a weight of 0.05.

\subsubsection{Datasets.} 
We train on the Adobe240 dataset\cite{adobe240}, which contains 133 videos at 720P resolution. During training, nine consecutive frames are selected, with the 1st and 9th frames as inputs and three randomly sampled intermediate frames as ground truth. For evaluation, we use the Vid4\cite{vid4}, Adobe240\cite{adobe240}, and GoPro datasets\cite{Nah_2017_CVPR}. The default spatial scale is $\times 4$. Unless otherwise specified, for Vid4, the temporal scale is $\times 2$ (center-frame interpolation). For GoPro-Average and Adobe-Average, the temporal scale is $\times 8$ (multi-frame interpolation). For GoPro-Center and Adobe-Center, only the 1st, 4th, and 9th frames are evaluated.

\subsubsection{Baselines and Metrics.} 
We compare against two categories of baselines. For F-STVSR, we include two-stage methods combining VFI models (SuperSloMo\cite{superslomo}, QVI\cite{qvi_nips19}, DAIN\cite{DAIN}) with VSR models (Bicubic, EDVR\cite{wang2019edvr}, BasicVSR\cite{chan2021basicvsr,chan2022basicvsrpp}), and one-stage methods (ZoomingSloMo\cite{xiang2020zooming}, TMNet\cite{xu2021temporal}). For C-STVSR, we include two-stage methods combining AISR models (LIIF\cite{liif}, LTE\cite{lte}) with VFI models (RIFE\cite{huang2022rife}, EMA-VFI\cite{zhang2023extracting}), and one-stage methods (VideoINR\cite{chen2022vinr}, MoTIF\cite{chen2023motif}, BF-STVSR\cite{kim2025bf}). All methods are evaluated using PSNR and SSIM on the Y channel.

\begin{figure}[htbp]
    \centering
    \includegraphics[width=\columnwidth]{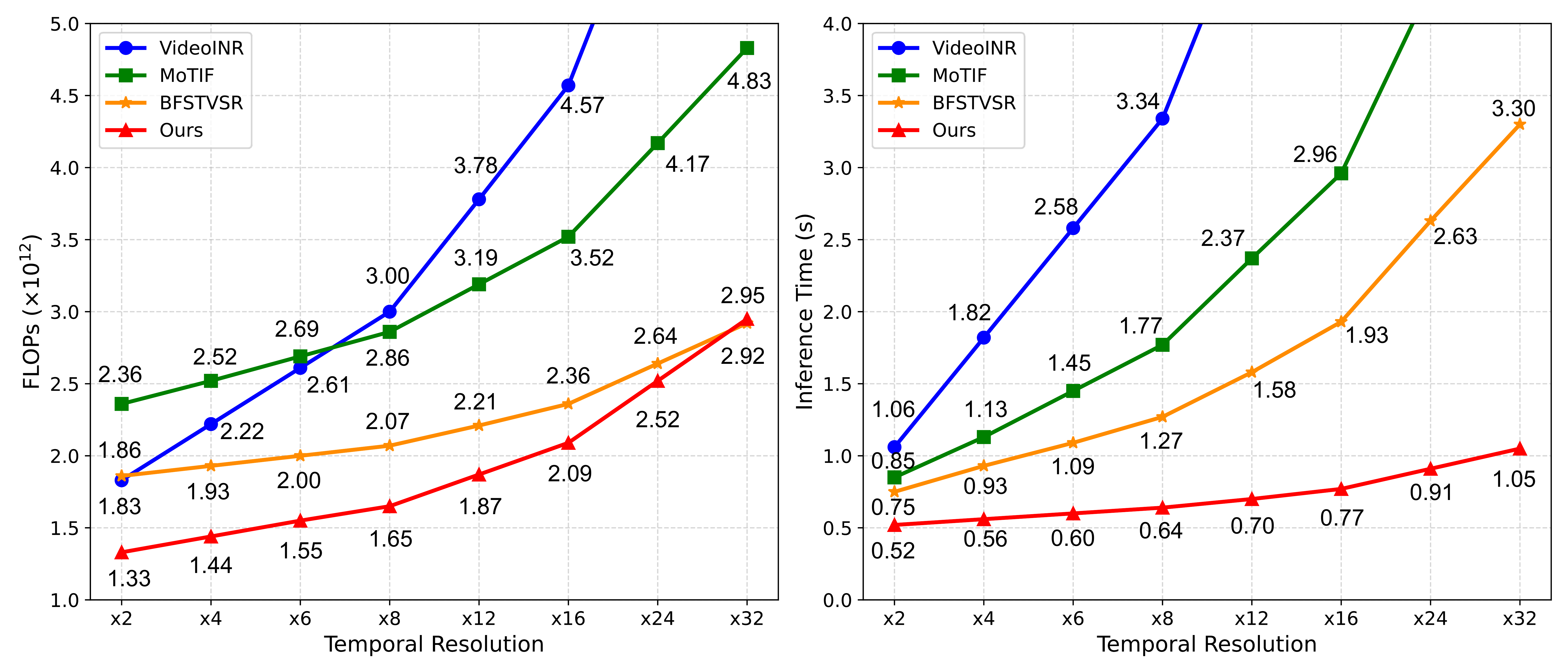}
    \caption{Computational cost (FLOPs, left) and inference time (right) comparison at $1280 \times 720$ resolution with $\times 4$ spatial upsampling across varying temporal scales ($\times 2$--$\times 32$).}
    \Description{Computational cost (left) and inference time (right) comparison on the spatial resolution of $1280 \times 720$ with different temporal scale. All frames are spatially interpolated by a factor of $\times 4$.}
    \label{fig:flops_time}
\end{figure}

\subsection{Quantitative Results}

Table~\ref{tab:t1} presents quantitative comparisons under the F-STVSR evaluation protocol on Vid4\cite{vid4}, GoPro\cite{Nah_2017_CVPR}, and Adobe240\cite{adobe240}. Among all methods, VideoINR\cite{chen2022vinr}, MoTIF\cite{chen2023motif}, BF-STVSR\cite{kim2025bf}, and our GS-STVSR are trained for C-STVSR, while the remaining methods are trained specifically for F-STVSR (results reproduced from BF-STVSR). Vid4, GoPro-Center, and Adobe-Center correspond to center-frame interpolation, whereas GoPro-Average and Adobe-Average evaluate multi-frame interpolation across all 8 intermediate frames. Our method achieves the best PSNR and SSIM across all five evaluation settings, consistently surpassing the previous state-of-the-art BF-STVSR. It is worth noting that our model also uses fewer parameters (12.67\,M vs.\ 13.47\,M) while delivering superior performance. Table~\ref{tab:t2} further evaluates C-STVSR performance under out-of-distribution spatiotemporal scales on the GoPro dataset. All one-stage C-STVSR methods are re-evaluated with identical settings, while two-stage results are reproduced from BF-STVSR. Our model attains the highest PSNR/SSIM in nearly all configurations, with the sole exception of the $\times 16$ temporal / $\times 4$ spatial setting where EMA-VFI+LIIF performs marginally better. Notably, the performance advantage of our method becomes more pronounced as the spatiotemporal scales increase, indicating strong generalization capability to challenging continuous upsampling scenarios not encountered during training.

\subsection{Qualitative Results}

Fig.~\ref{fig:visual} presents visual comparisons with VideoINR\cite{chen2022vinr}, MoTIF\cite{chen2023motif}, and BF-STVSR\cite{kim2025bf} at both in-distribution ($\times 8$) and out-of-distribution ($\times 6$) temporal scales. For the in-distribution scale ($\times 8$), we present a scenario featuring a fast-moving motorcycle. Our model successfully and accurately predicts the position of the motorcycle rider in any intermediate frame; although minor local deformations exist, the majority of the information is well-preserved. In contrast, competing methods fail to accurately reconstruct the content and produce noticeable dot-matrix-like artifacts, which severely degrade the visual quality. Consequently, our method yields much cleaner temporal transitions. For the out-of-distribution scale ($\times 6$), we present a scene involving mutually moving and occluding objects. As highlighted in the edge regions adjacent to the house and utility pole, our model remains stable without distorting foreground and background information due to occlusion. In comparison, other models exhibit distortion phenomena with obvious dot-matrix-like pixel artifacts in the overlapping areas, impairing visual perception, whereas our method produces significantly more natural and coherent interpolation. These visual comparisons confirm that GS-STVSR is capable of delivering natural motion interpolation and effectively preserving fine-grained textures across diverse temporal scales.

\subsection{Computational Cost and Latency}
\noindent Fig.~\ref{fig:flops_time} compares FLOPs and inference time across temporal scales from $\times 2$ to $\times 32$, measured on an NVIDIA Tesla V100 GPU at $1280 \times 720$ resolution with $\times 4$ spatial upsampling (inference time averaged over 100 runs). We use the \textbf{fvcore} library\footnote{https://github.com/facebookresearch/fvcore} to measure FLOPs. Our method consistently achieves the lowest computational cost and fastest inference across all temporal scales. At conventional scales ($\times 2$--$\times 8$), our model substantially outperforms BF-STVSR in both FLOPs and wall-clock time, with inference latency remaining nearly constant at a very low level. Remarkably, our inference time at the $\times 16$ temporal scale is even comparable to that of BF-STVSR at $\times 2$. Even at the extreme $\times 32$ scale, our method maintains over $3\times$ speedup compared to BF-STVSR. This near-constant latency arises from our architecture design: the encoder features and optical flows are extracted once and shared across all intermediate time steps, while the subsequent Gaussian parameter prediction and 2D-GS rasterization per frame incur only minimal overhead, demonstrating strong practical applicability for real-time video processing.

\begin{table}[htbp]
    \centering
    \caption{Ablation study of key modules. \textbf{CRA}: Covariance Resampling Alignment, \textbf{AOW}: Adaptive Offset Window. Best results are \textbf{bolded}.}
    \label{tab:ablation}
    \renewcommand{\arraystretch}{1.3} % 增加行高, 看起来更舒展
    \begin{tabular}{cccccc} 
        \toprule
        % 使用更清晰的表头
        \textbf{Index} & \textbf{CRA} & \textbf{AOW} & GoPro-Average $\uparrow$ & Adobe-Average $\uparrow$ \\
        \midrule
        (a) & \ding{51}  & \ding{51}  & \textbf{30.35 / 0.8817} & \textbf{30.35 / 0.8827} \\ 
        (b) & \ding{51}  &  \ding{55} & 30.26 / 0.8808          & 30.29 / 0.8821          \\
        (c) & \ding{55}  & \ding{51}  & 30.28 / 0.8808          & 30.23 / 0.8818          \\
        (d) & \ding{55}  &  \ding{55} & 30.17 / 0.8801          & 30.15 / 0.8809          \\
        \bottomrule
    \end{tabular}
\end{table}

\subsection{Ablation Study}

Table~\ref{tab:ablation} reports ablation results for the two proposed modules: The full model, equipped with both the CPA and AOW modules, achieves the best performance across both benchmarks. Removing either module individually or both simultaneously leads to a noticeable degradation in performance. It is worth noting that, under identical training configurations, both model variants lacking the CPA module exhibit a performance drop after $400\text{K}$ iterations. Notably, the variant with only the CPA module removed suffers from severe training collapse. Consequently, we report the optimal metrics obtained at the $400\text{K}$-iteration mark for these specific models. This phenomenon validates our hypothesis: while inter-frame covariance exhibits exceptionally strong temporal correlation and can be easily captured by a lightweight module, the unconstrained transition of these parameters is highly susceptible to covariance drift, which severely destabilizes the training process.

\begin{figure}[t]
    \centering
    \includegraphics[width=\columnwidth]{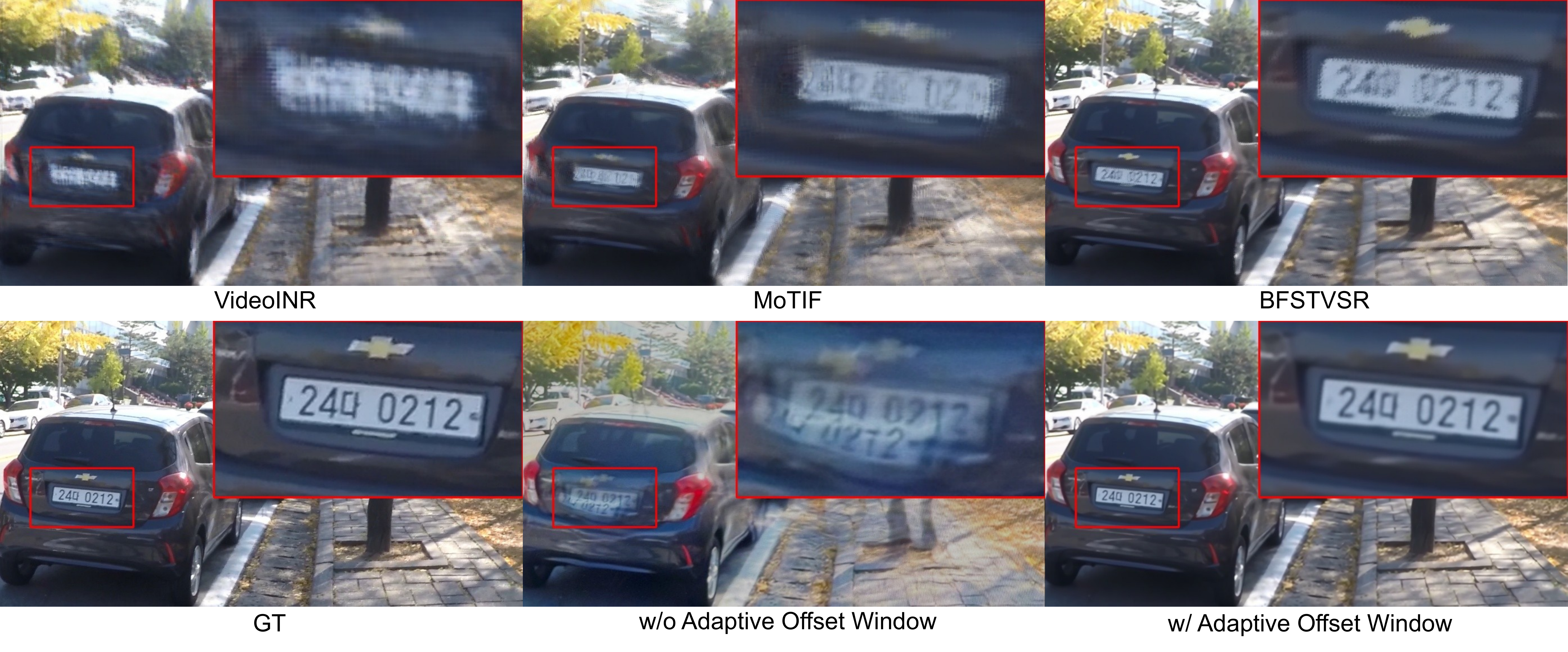}
    \caption{Visual comparison at $t = 0.5$ on a GoPro test scene with large camera shake ($\times 4$ spatial, $\times 8$ temporal).}
    \Description{Visual results at $t = 0.5$ for a specific scene from GoPro dataset, with a spatial scale of $\times 4$ and a temporal scale of $\times 8$}
    \label{fig:motion}
\end{figure}

To further illustrate the practical effect of AOW, we visualize a representative scene from the GoPro test set characterized by significant camera shake in Fig.~\ref{fig:motion}. Without AOW, the constrained offset range results in severe multi-frame overlapping artifacts, which are effectively eliminated upon enabling AOW. Furthermore, we compare our proposed method with other C-STVSR approaches in the same scene. Under such large-scale motions, competing methods tend to suffer from pronounced aliasing effects and fail to faithfully reconstruct the high-frequency textures present in the ground truth. In contrast, our approach maintains robust performance and delivers sharper visual results.

\section{Limitations and Future Work}

While GS-STVSR achieves strong efficiency and reconstruction quality, several limitations remain. First, the linear motion assumption for optical flow interpolation may introduce alignment errors under complex non-linear motions (e.g., rotation or acceleration). Second, our framework's performance is upper-bounded by the pre-trained optical flow extractor, making it susceptible to errors in challenging scenarios like motion boundaries or textureless regions. Third, the pre-defined and fixed CPB may restrict representational capacity for diverse video content. Future work will explore learned non-linear motion models, end-to-end joint optimization of optical flow and Gaussian parameters, and scene-adaptive CPBs.

\section{Conclusion}

We propose GS-STVSR, the first 2D Gaussian Splatting framework for continuous space-time video super-resolution. By representing frames as continuously evolving 2D Gaussians, it bypasses the computational bottlenecks of INR-based methods. Based on the strong temporal stability of covariance, we design an ultra-lightweight prediction module. Together with optical flow guidance and adaptive offset windows (AOW), this ensures accurate Gaussian evolution under varying motions, while our Covariance Resampling Alignment (CRA) prevents covariance drift. Experiments show that GS-STVSR achieves state-of-the-art reconstruction quality with fewer parameters and over $3\times$ inference speedup at extreme temporal scales, proving Gaussian representations are a highly efficient paradigm for video enhancement.

\clearpage

\bibliographystyle{ACM-Reference-Format}
\bibliography{main}

\clearpage

\appendix

\section{Details of Covariance Experiment}

\begin{figure}[htbp]
    \centering
    \includegraphics[width=\columnwidth]{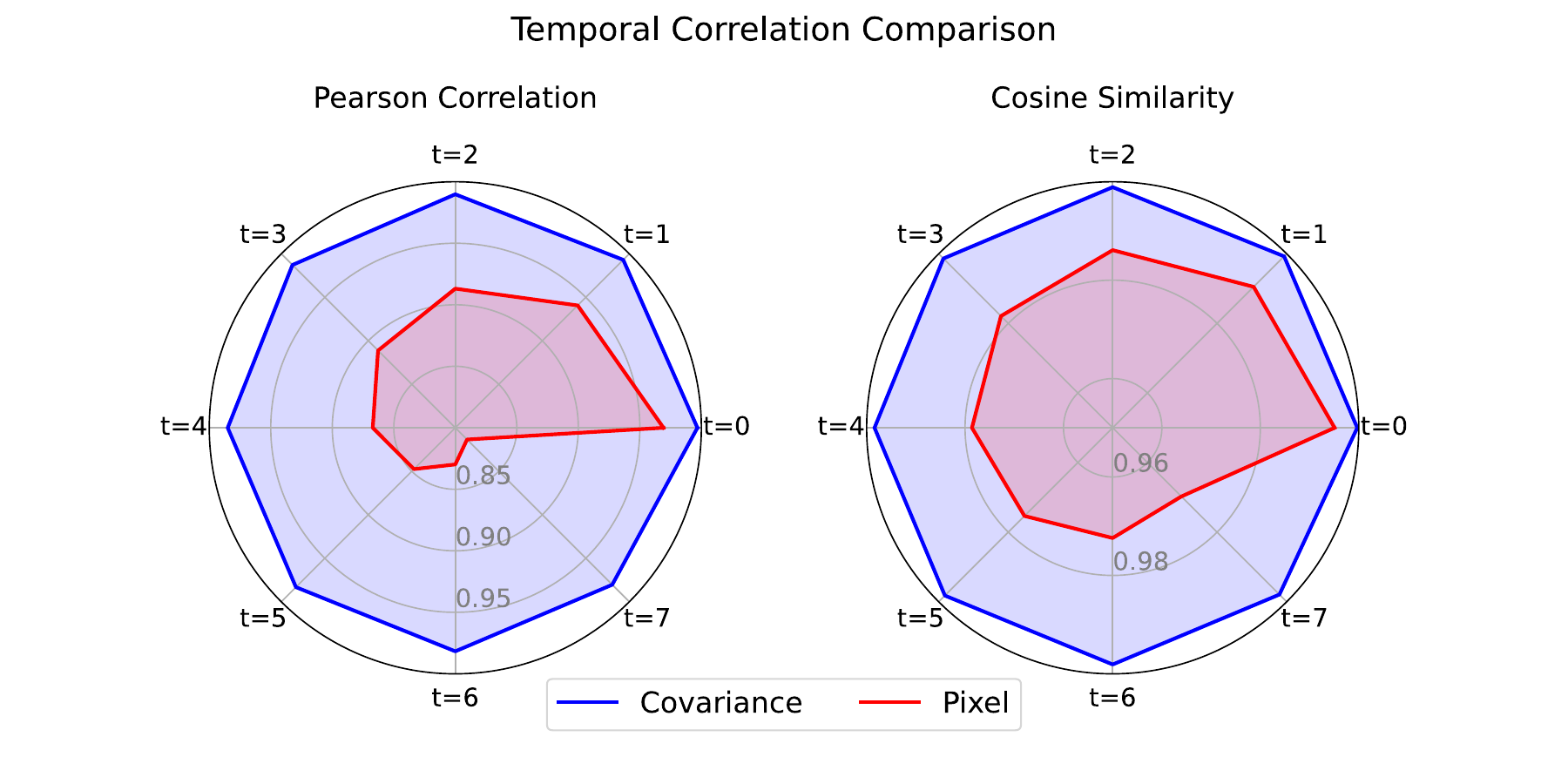}
    \caption{Comparison of temporal correlation between pixel-level and covariance-based representations. \textmd{The radar charts evaluate the correlation across continuous time intervals ($t=0$ to $7$) using Pearson Correlation (left) and Cosine Similarity (right). Gaussian covariance (blue) exhibits significantly higher and more stable temporal consistency than raw pixels (red).}}
    \Description{Comparison of temporal correlation between pixel-level and covariance-based representations. The radar charts evaluate the correlation across continuous time intervals ($t=0$ to $7$) using Pearson Correlation (left) and Cosine Similarity (right). Gaussian covariance (blue) exhibits significantly higher and more stable temporal consistency than raw pixels (red).}
    \label{fig:temporal}
\end{figure}

To elucidate our motivation, we first investigate the temporal stability of various scene representations. Traditional STVSR methods rely heavily on pixel-level temporal alignment, which often suffers from severe performance degradation under complex or large-scale motions. In contrast, modeling scene dynamics with 2D Gaussian functions provides a fundamentally more robust alternative. As illustrated in Fig.~\ref{fig:temporal}, we conducted a proof-of-concept experiment to compare the temporal correlation of raw pixel values against Gaussian covariance parameters across different time intervals (from $t = 0$ to $t = 7$). We employed Pearson correlation and cosine similarity as our evaluation metrics. The radar chart clearly illustrates this discrepancy: the pixel-level correlation (represented by the red polygon) decays rapidly as the time interval increases, indicating its high sensitivity to temporal variations. Conversely, the covariance parameters (represented by the blue polygon) exhibit exceptional stability, forming a nearly uniform and maximal area across all time steps. This observation demonstrates, both visually and quantitatively, that Gaussian covariance possesses superior robustness in handling temporal variations. Consequently, leveraging these highly stable covariance parameters forms the core foundation of our proposed C-STVSR framework, enabling precise motion modeling and high-fidelity feature aggregation.

The aforementioned experiments were evaluated on the GoPro\cite{Nah_2017_CVPR} dataset using \cite{ptg}. We also conducted identical experiments on the Adobe240\cite{adobe240} dataset, which yielded consistent results; thus, we omit the redundant visualizations. Furthermore, we concurrently analyzed the temporal stability of the position and color attributes. The color information exhibits trends and similarities highly consistent with those in the pixel domain. Although the position information demonstrates lower similarity compared to the covariance parameters, it still maintains commendable stability, presenting a promising avenue for our future work.

\begin{figure*}[htbp]
    \centering
    % --- 第一个子图 (左图: 摄像机快速变焦) ---
    \begin{subfigure}[b]{0.48\textwidth}
        \centering
        \includegraphics[width=\textwidth]{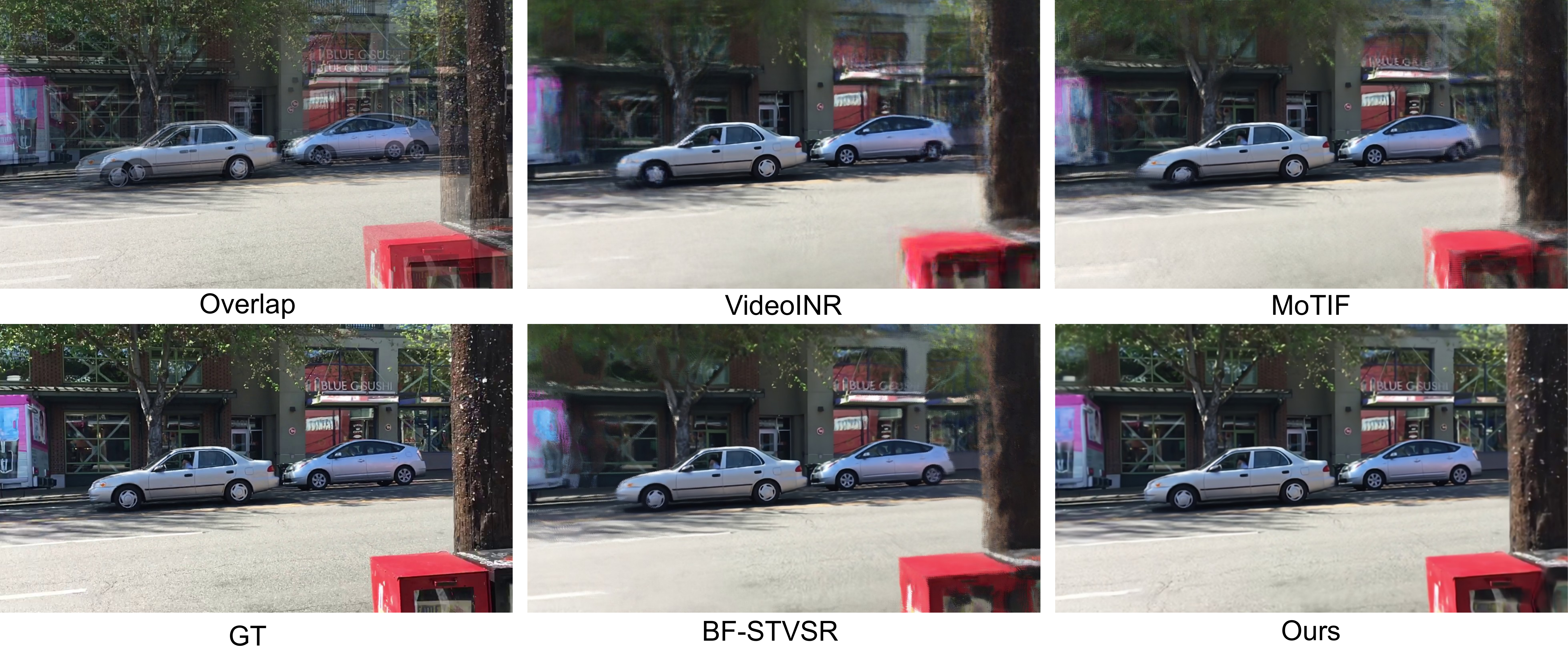}
        \caption{Rapid Camera Zooming}
        \label{fig:camera_zoom}
    \end{subfigure}
    \hfill
    % --- 第二个子图 (右图: 摄像机快速平移/运动) ---
    \begin{subfigure}[b]{0.48\textwidth}
        \centering
        \includegraphics[width=\textwidth]{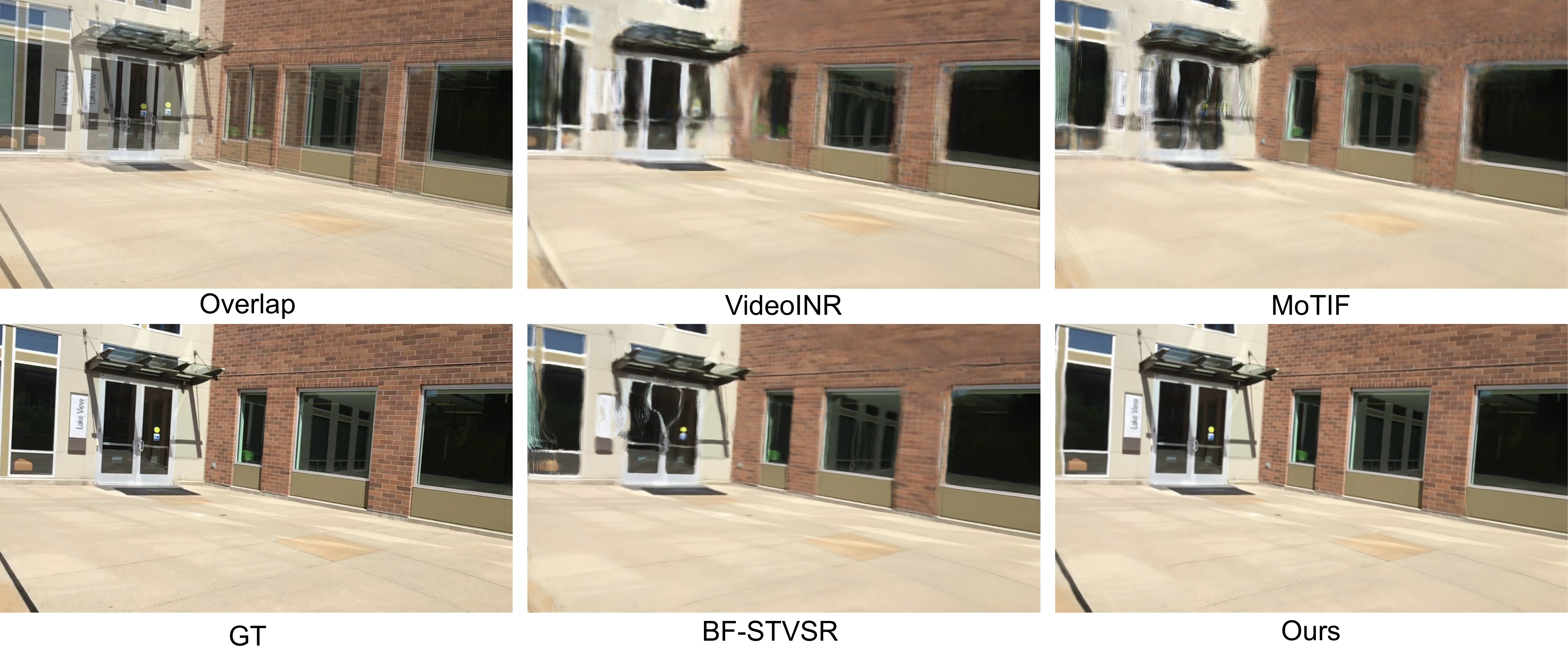}
        \caption{Fast Camera Translation}
        \label{fig:camera_translation}
    \end{subfigure}
    % --- 总图注 ---
    \caption{Qualitative comparison of different C-STVSR methods under large-scale global camera motions. \textmd{(\subref{fig:camera_zoom}) Visual results in a scenario with rapid camera zooming, where severe global scale changes occur. (\subref{fig:camera_translation}) Visual results under fast camera translation, causing massive global pixel shifts. Best viewed zoomed in.}}
    \label{fig:global_motion}
\end{figure*}

\section{Additional Quantitative Results}

\begin{table}[htbp]
\centering
\caption{Performance comparison on the C-STVSR baselines for out-of-distribution scale on Adobe240\cite{adobe240} dataset.\textmd{Results are evaluated using PSNR (dB) and SSIM metrics in Y channels. All frames are interpolated by a scaling factor specified on the table and metrics calculated across all interpolated frames. \textcolor{red}{Red} and \textcolor{blue}{blue} indicate the best and the second best performance, respectively.}}
\label{tab:otf-adobe} 
\resizebox{0.48\textwidth}{!}{
\begin{tabular}{cccccc}
\toprule
Temporal & Spatial & \multirow{2}{*}{VideoINR\cite{chen2022vinr}} & \multirow{2}{*}{MoTIF\cite{chen2023motif}} & \multirow{2}{*}{BF-STVSR\cite{kim2025bf}} & \multirow{2}{*}{Ours} \\
Scale & Scale & & & & \\
\midrule
$\times 8$ & $\times 4$ & 29.27 / 0.8651 & 29.82 / 0.8750 & \textcolor{blue}{30.12 / 0.8808} & \textcolor{red}{30.35 / 0.8827} \\
\addlinespace
\multirow{3}{*}{$\times 6$} & $\times 4$ & 30.08 / 0.8842 & 30.79 / 0.8926 & \textcolor{blue}{31.05} / \textcolor{red}{0.8973} & \textcolor{red}{31.09} / \textcolor{blue}{0.8942} \\
& $\times 6$ & 27.52 / 0.8057 & 28.20 / 0.8169 & \textcolor{blue}{28.40} / \textcolor{red}{0.8220} & \textcolor{red}{28.49} / \textcolor{blue}{0.8217} \\
& $\times 12$ & 23.90 / 0.6733 & 24.80 / 0.6904 & \textcolor{blue}{24.84 / 0.6906} & \textcolor{red}{25.03 / 0.6964} \\
\addlinespace
\multirow{3}{*}{$\times 12$} & $\times 4$ & 27.25 / 0.8203 & 27.67 / 0.8274 & \textcolor{blue}{28.04 / 0.8362} & \textcolor{red}{28.30 / 0.8398} \\
& $\times 6$ & 25.89 / 0.7658 & 26.32 / 0.7756 & \textcolor{blue}{26.74 / 0.7861} & \textcolor{red}{26.90 / 0.7873} \\
& $\times 12$ & 23.25 / 0.6594 & 24.04 / 0.6761 & \textcolor{blue}{24.31 / 0.6802} & \textcolor{red}{24.48 / 0.6854} \\
\addlinespace
\multirow{3}{*}{$\times 16$} & $\times 4$ & 25.77 / 0.7795 & 26.07 / 0.7844 & \textcolor{blue}{26.46 / 0.7947} & \textcolor{red}{26.71 / 0.7998} \\
& $\times 6$ & 24.87 / 0.7381 & 25.15 / 0.7452 & \textcolor{blue}{25.64 / 0.7577} & \textcolor{red}{25.80 / 0.7603} \\
& $\times 12$ & 22.78 / 0.6493 & 23.42 / 0.6637 & \textcolor{blue}{23.82 / 0.6698} & \textcolor{red}{23.99 / 0.6754} \\
\bottomrule
\end{tabular}
}
\end{table}

\begin{table}[htbp]
\centering
\caption{Performance comparison on the Arbitrary Video Super-Resolution (AVSR) baselines for different spatial scales on Adobe240\cite{adobe240} and GoPro\cite{Nah_2017_CVPR} datasets.\textmd{Results are evaluated using PSNR (dB) and SSIM metrics in Y channels. \textcolor{red}{Red} and \textcolor{blue}{blue} indicate the best and the second best performance, respectively.}}
\label{tab:avsr} 
\resizebox{0.48\textwidth}{!}{
\begin{tabular}{cccccc}
\toprule
Dataset & Spatial Scale & VideoINR\cite{chen2022vinr} & MoTIF\cite{chen2023motif} & BF-STVSR\cite{kim2025bf} & Ours \\
\midrule
\multirow{4}{*}{GoPro\cite{Nah_2017_CVPR}} 
& $\times 4$ & 32.88 / 0.9297 & 34.13 / 0.9368 & \textcolor{blue}{34.29 / 0.9381} & \textcolor{red}{34.57 / 0.9404} \\
& $\times 6$ & 29.52 / 0.8574 & 30.28 / 0.8649 & \textcolor{blue}{30.35 / 0.8660} & \textcolor{red}{30.41 / 0.8679} \\
& $\times 8$ & 26.70 / 0.7901 & \textcolor{blue}{27.13 / 0.7993} & 27.00 / 0.7976 & \textcolor{red}{27.36 / 0.8036} \\
& $\times 12$ & 24.84 / 0.7164 & \textcolor{blue}{25.59 / 0.7275} & 25.50 / 0.7221 & \textcolor{red}{25.71 / 0.7324} \\
\addlinespace
\multirow{4}{*}{Adobe240\cite{adobe240}} 
& $\times 4$ & 31.44 / 0.9029 & \textcolor{blue}{32.50 / 0.9109} & \textcolor{blue}{32.62 / 0.9123} & \textcolor{red}{33.02 / 0.9167} \\
& $\times 6$ & 28.26 / 0.8171 & 28.92 / 0.8247 & \textcolor{blue}{28.99 / 0.8267} & \textcolor{red}{29.14 / 0.8293} \\
& $\times 8$ & 25.73 / 0.7466 & \textcolor{blue}{26.17 / 0.7562} & 26.04 / 0.7557 & \textcolor{red}{26.45 / 0.7620} \\
& $\times 12$ & 24.08 / 0.6745 & \textcolor{blue}{24.80 / 0.6868} & 24.73 / 0.6837 & \textcolor{red}{24.96 / 0.6927} \\
\bottomrule
\end{tabular}
}
\end{table}

\subsection{Continuous Spatio-Temporal Super-Resolution}
To further evaluate the robustness and generalization ability of our proposed method, we conduct extensive experiments on out-of-distribution (OOD) spatio-temporal scales. As shown in Table~\ref{tab:otf-adobe}, we additionally evaluate our model on the Adobe240 dataset under extreme and arbitrary scaling factors. Under all challenging configurations, our method consistently outperforms state-of-the-art baselines, including VideoINR, MoTIF, and BF-STVSR. The superior PSNR and SSIM scores indicate that modeling spatio-temporal dynamics via 2D Gaussians provides a more continuous and robust representation, further demonstrating the enhanced generalization capability of our model. This effectively mitigates the severe performance degradation typically observed in traditional grid-based or pixel-level interpolation methods when pushed to extreme continuous scales.

\subsection{Arbitrary Video Super-Resolution} 
In addition to the joint spatio-temporal evaluation, we also investigate the capability for purely spatial arbitrary video super-resolution. Table~\ref{tab:avsr} presents the quantitative results on the GoPro and Adobe240 datasets for spatial upsampling scales ranging from $4\times$ to $12\times$. Even when performing the AVSR task independently, our framework maintains a distinct advantage over existing methods. Notably, at the highly challenging $12\times$ spatial scale, our method surpasses the second-best performing model by a significant margin. This consistent superiority validates the effectiveness of our feature aggregation and rendering mechanisms in reconstructing high-frequency spatial details at arbitrary resolutions. However, due to certain inherent limitations of 2D-GS, our model currently cannot perform arbitrary-scale frame interpolation under a $2\times$ spatial super-resolution condition. We attribute this to our representation strategy, where the ratio of the number of initialized Gaussian kernels to the original pixels is set to 1:4. This specific density ratio leads to erroneous rendering results when the spatial upsampling factor falls below $2\times$.

\section{Additional Qualitative Results}

\begin{figure*}[htbp]
    \centering
    \includegraphics[width=0.9\textwidth]{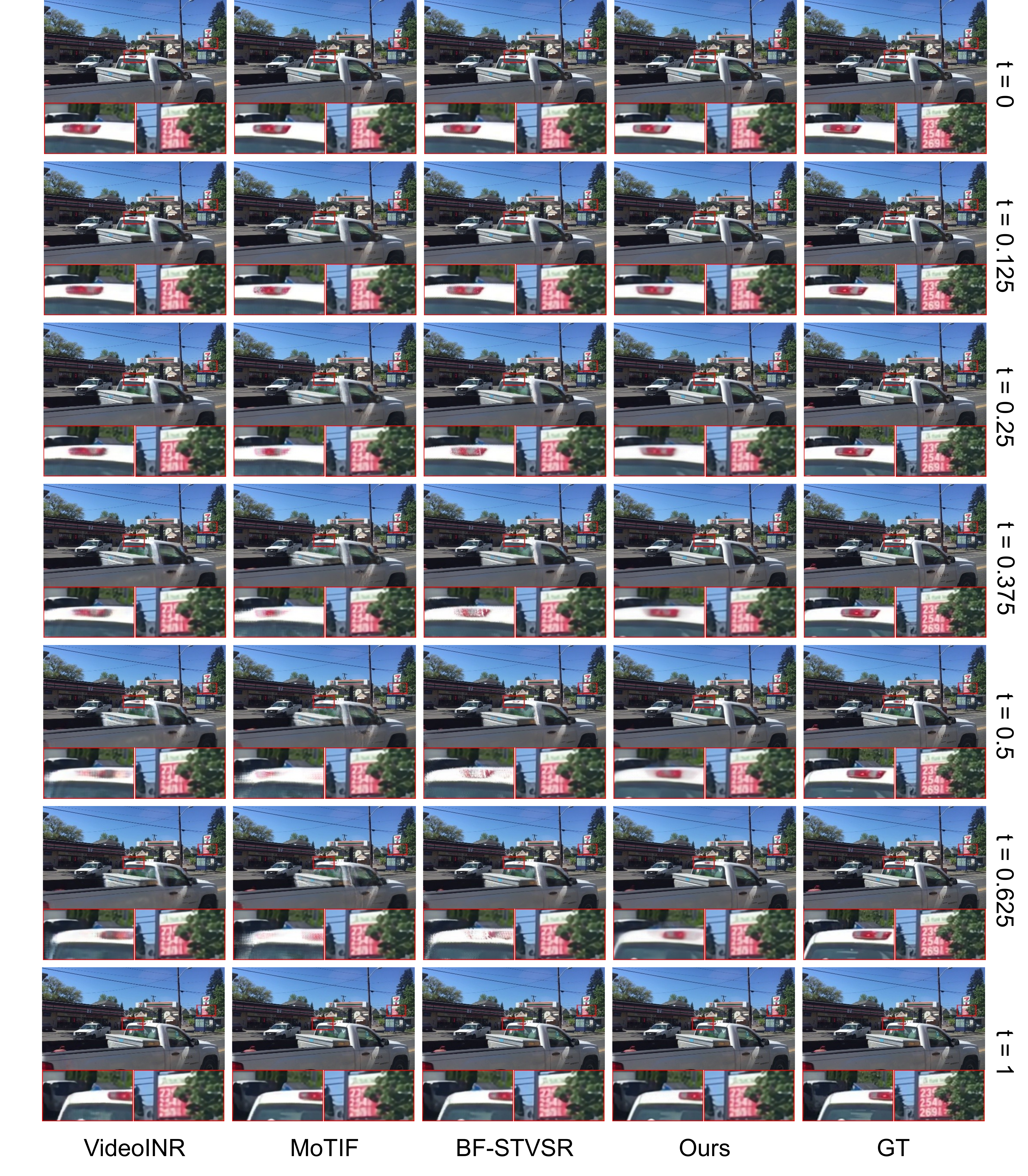}
    \caption{Qualitative comparisons of different C-STVSR methods on in-distribution time scale. The spatial scaling factor is set to $\times 4$ and the temporal factor to $\times 8$. Best viewed zoomed in.}
    \Description{Qualitative comparisons of different C-STVSR methods on in-distribution time scale. The spatial scaling factor is set to $\times 4$ and the temporal factor to $\times 8$. Best viewed zoomed in.}
    \label{fig:sx4_tx8}
\end{figure*}

\begin{figure*}[htbp]
    \centering
    \includegraphics[width=0.9\textwidth]{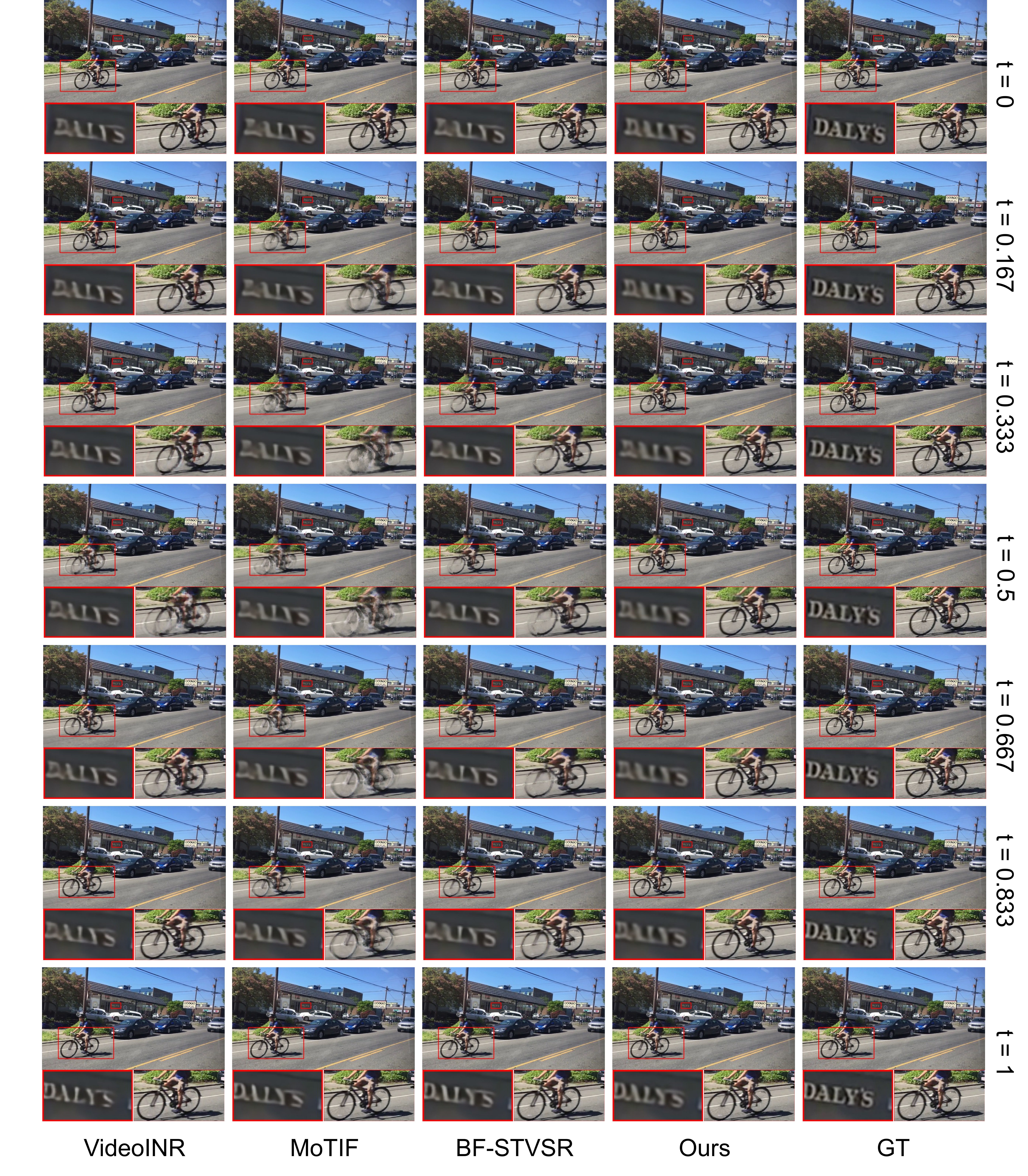}
    \caption{Qualitative comparisons of different C-STVSR methods on out-of-distribution time scale. The spatial scaling factor is set to $\times 4$ and the temporal factor to $\times 6$. Best viewed zoomed in.}
    \Description{Qualitative comparisons of different C-STVSR methods on out-of-distribution time scale. The spatial scaling factor is set to $\times 4$ and the temporal factor to $\times 6$. Best viewed zoomed in.}
    \label{fig:sx4_tx6}
\end{figure*}

\begin{figure*}[htbp]
    \centering
    \includegraphics[width=0.85\textwidth]{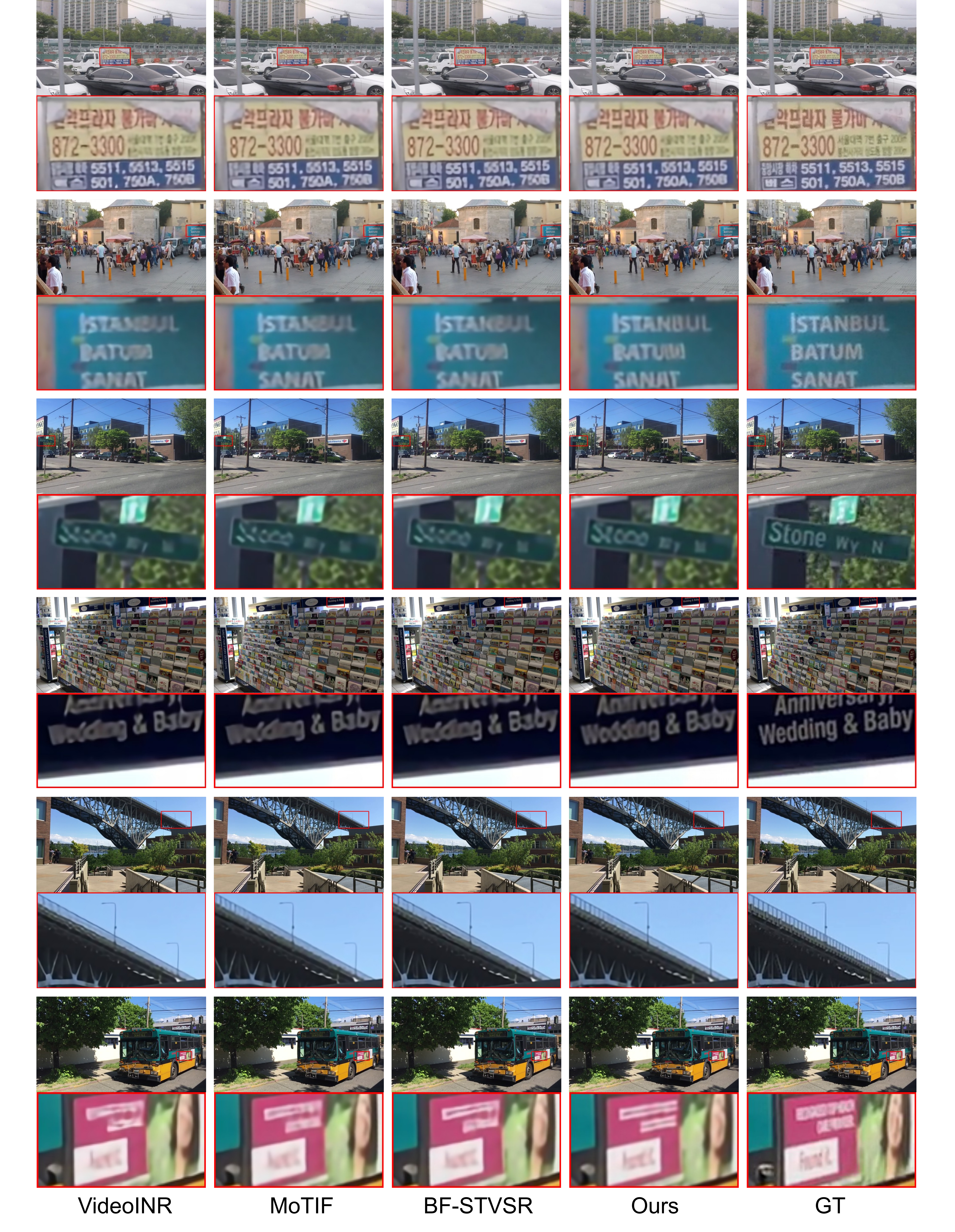}
    \caption{Qualitative comparison of different C-STVSR methods in the VSR task. The spatial scaling factor is set to $\times 4$. Best viewed zoomed in.}
    \Description{Qualitative comparison of different C-STVSR methods in the VSR task. The spatial scaling factor is set to $\times 4$. Best viewed zoomed in.}
    \label{fig:sx4_tx1}
\end{figure*}

In this section, we provide qualitative comparisons between our proposed method and state-of-the-art C-STVSR baselines under various scaling configurations. First, Fig.~\ref{fig:sx4_tx8} visualizes the super-resolution results on an \textbf{in-distribution temporal scale}, specifically evaluating the configuration with a spatial scaling factor of $\times 4$ and a temporal factor of $\times 8$. Second, to demonstrate the generalization capability of our model, Fig.~\ref{fig:sx4_tx6} presents qualitative comparisons on an \textbf{out-of-distribution temporal scale}. In this scenario, the models are evaluated using a spatial scaling factor of $\times 4$ and a temporal factor of $\times 6$. Finally, Fig.~\ref{fig:sx4_tx1} illustrates the visual performance of different methods restricted to the standard \textbf{Video Super-Resolution (VSR) task}, where only spatial upsampling (with a factor of $\times 4$) is performed without temporal frame interpolation. Across all scenarios, our method consistently synthesizes visually pleasing frames with sharper details, clearer structural boundaries, and fewer blurring artifacts compared to the existing baselines.

To further demonstrate the robustness of our proposed framework, we provide additional qualitative comparisons under scenarios dominated by large-scale global camera motions in Fig.~\ref{fig:global_motion}. These challenging cases, involving rapid zooming and fast translation, test the model's ability to handle massive, non-rigid pixel displacements across the entire frame. Specifically, Fig.~\ref{fig:global_motion}(a) visualizes a scene with rapid camera zooming. The severe global scale change causes previous C-STVSR baselines to produce noticeable ghosting artifacts and blurred edges. In contrast, our method effectively adapts to these variations, maintaining sharp structural boundaries. Furthermore, Fig.~\ref{fig:global_motion}(b) illustrates a scenario with fast camera translation. While massive global pixel shifts almost entirely ruin the interpolation results of other models (exhibiting severe pixelation and structural distortion), our model maintains remarkable stability. By correctly modeling the global trajectory, our framework synthesizes artifact-free frames that strictly preserve both foreground details and background contexts. These results demonstrate that our Gaussian-based representation is intrinsically more robust against extreme temporal variations than traditional pixel-based methods.

\section{Computational Efficiency}
\begin{table}[htbp]
\centering
\caption{Computational efficiency comparison for $270\text{p} \rightarrow 1080\text{p}$ video upscaling. \textmd{Results are evaluated using inference time (s), render time (s), and VRAM (GiB). \textcolor{red}{Red} and \textcolor{blue}{blue} indicate the best and the second best performance, respectively. Note that lower values indicate better efficiency.}}
\label{tab:efficiency} 
\resizebox{0.48\textwidth}{!}{
\begin{tabular}{ccccccc}
\toprule
Temporal & Spatial & \multirow{2}{*}{Metric} & \multirow{2}{*}{VideoINR\cite{chen2022vinr}} & \multirow{2}{*}{MoTIF\cite{chen2023motif}} & \multirow{2}{*}{BF-STVSR\cite{kim2025bf}} & \multirow{2}{*}{Ours} \\
Scale & Scale & & & & & \\
\midrule
\multirow{3}{*}{$\times 8$} & \multirow{3}{*}{$\times 4$} 
& Inference Time (s) $\downarrow$ & 7.89 & 4.3 & \textcolor{blue}{3.0} & \textcolor{red}{1.4} \\
& & Rendering Time (s) $\downarrow$ & 7.13 & 3.5 & \textcolor{blue}{2.2} & \textcolor{red}{0.4} \\
& & VRAM (GiB) $\downarrow$ & \textcolor{red}{13.8} & 32.0 & 29.3 & \textcolor{blue}{15.5} \\
\addlinespace
\multirow{3}{*}{$\times 4$} & \multirow{3}{*}{$\times 4$} 
& Inference Time (s) $\downarrow$ & 4.3 & 2.7 & \textcolor{blue}{2.1} & \textcolor{red}{1.3} \\
& & Rendering Time (s) $\downarrow$ & 3.7 & 1.997 & \textcolor{blue}{1.4} & \textcolor{red}{0.3} \\
& & VRAM (GiB) $\downarrow$ & \textcolor{red}{13.7} & 32.2 & 28.8 & \textcolor{blue}{15.3} \\
\bottomrule
\end{tabular}
}
\end{table}

A critical advantage of Gaussian-based representations lies in their computational efficiency, particularly benefiting from the highly optimized rasterization pipeline. To reflect practical, real-world application scenarios, we benchmark our model by upscaling videos from 270p ($480 \times 270$) to 1080p ($1920 \times 1080$) under $4\times$ and $8\times$ temporal configurations. We comprehensively evaluate three metrics: full inference time (the complete forward pass including the encoder), pure rendering time (excluding the encoder, which accurately represents the core C-STVSR rasterization process), and VRAM consumption. All time-related metrics are calculated as the average of 100 continuous inference iterations to ensure robust measurement. As detailed in Table~\ref{tab:efficiency}, our method achieves a remarkable reduction in processing time. For instance, under the $8\times$ temporal configuration, our pure rendering time is merely $0.4$ seconds, which is over $5$ times faster than the recent BF-STVSR framework. Furthermore, while maintaining this exceptionally high speed and state-of-the-art visual quality, our framework requires significantly less VRAM (15.5 GiB) compared to MoTIF (32.0 GiB) and BF-STVSR (29.3 GiB). Although VideoINR consumes slightly less VRAM, it does so at the cost of substantially lower rendering speed and inferior synthesis quality. Overall, our method achieves an optimal balance between performance and computational efficiency, paving the way for practical, high-resolution continuous video interpolation.

\end{document}